\documentclass[preprint,12pt]{elsarticle}

\usepackage[utf8]{inputenc}
\usepackage{amsmath,amsfonts,amssymb,amsthm,bm,mathtools}
\usepackage{xcolor}
\usepackage{hyperref}
\usepackage{siunitx}
\usepackage{graphicx}
\usepackage{float}
\usepackage{subcaption}
\usepackage{tabularray}
\usepackage{cleveref}

\crefname{figure}{Fig.}{Figs.}
\Crefname{figure}{Fig.}{Figs.}
\crefname{table}{Tab.}{Tabs.}
\Crefname{table}{Tab.}{Tabs.}
\crefname{equation}{Eq.}{Eqs.}
\Crefname{equation}{Eq.}{Eqs.}
\crefname{section}{Sec.}{Secs.}
\Crefname{section}{Sec.}{Secs.}

\UseTblrLibrary{booktabs}

\DeclareMathOperator*{\argmax}{arg\,max}

\DeclarePairedDelimiterX{\infdivx}[2]{(}{)}{%
  #1\;\delimsize\|\;#2%
}

\begin{document}
\begin{frontmatter}

\title{Generative AI-enhanced Probabilistic Multi-Fidelity Surrogate Modeling Via Transfer Learning}
\author[PNNL]{Jice Zeng}

\author[PNNL]{David Barajas-Solano\corref{label1}}
\ead{David.Barajas-Solano@pnnl.gov}
\cortext[label1]{Correspondence authors.}

\author[WuHan]{Hui Chen}

\affiliation[PNNL]{organization={Pacific Northwest National Laboratory},%
city={Richland},
postcode={99352},
state={WA, USA}}
\affiliation[WuHan]{organization={School of Civil Engineering Architecture, Wuhan University of Technology},%
    city={WuHan, China},%
    postcode={430070}}

\begin{abstract}
The performance of machine learning surrogates is critically dependent on data quality and quantity. This presents a major challenge, as high-fidelity (HF) data is often scarce and computationally expensive to acquire, while low-fidelity (LF) data is abundant but less accurate. To address this data-scarcity problem, we propose a probabilistic multi-fidelity surrogate modeling framework that integrates transfer learning with generative modeling. We employ a normalizing flow (NF) generative model as the backbone, which is trained in two phases: (i) the NF is first pretrained on a large LF dataset to learn a probabilistic forward model; (ii) the pretrained model is then fine-tuned on a small HF dataset, allowing it to correct for LF--HF discrepancies via knowledge transfer. To relax the dimension-preserving constraint of standard bijective NFs, we integrate surjective (dimension-reducing) layers with standard coupling blocks. This architecture enables learned dimension reduction while preserving the ability to train with exact likelihoods.
The resulting surrogate provides fast probabilistic predictions with quantified uncertainty and significantly outperforms LF-only baselines while using fewer HF evaluations. We validate the approach for two benchmark systems: a rail-sleeper-ballast and a reinforced concrete slab. For both applications, we combine many coarse-mesh (LF) simulations with a limited set of fine-mesh (HF) simulations. The proposed model achieves probabilistic predictions with HF accuracy, demonstrating a practical path toward data-efficient, generative AI-driven surrogates for complex engineering systems.
\end{abstract}


\end{frontmatter}


\section{Introduction}\label{sec:intro}

High-fidelity (HF) computer modeling using discretization schemes such as the finite elements (FE) method provides a rigorous framework for analyzing and predicting the behavior of complex engineering systems. Nevertheless, while powerful, querying such HF FE models is often computationally expensive, and this cost becomes prohibitive for tasks requiring thousands of queries, such as uncertainty quantification \cite{bae2004epistemic}, reliability analysis \cite{zhao2018reliability,zeng2026deep}, probabilistic model inversion \cite{zeng2025solving,zeng2023probabilistic,zong2025randomized}, and Bayesian model updating \cite{zeng2023bayesian,wan2016stochastic}, among others. Data-driven deep learning-based surrogates have emerged as an effective approach to accelerating these computations. However, these models face their own critical challenge, namely that the performance of deep learning methods is highly dependent on the quality and quantity of available training data \cite{kudela2022recent,samadian2025application}.

Training data for surrogate modeling is typically generated from model queries, which can often be performed at different levels of fidelity. Low-fidelity (LF) models, which may employ coarse meshes or simplified physics, are computationally inexpensive and can generate large training datasets quickly. However, these simplifications often introduce significant bias, limiting the surrogate's predictive accuracy.
HF models, on the other hand, introduce less bias, but generating sufficiently large training datasets with such models is computationally demanding, leading to a cost-versus-bias trade-off: relying solely on LF data leads to inaccurate surrogates, while relying on HF data is computationally intractable. This challenge motivates the development of multi-fidelity surrogate modeling, a strategy that fuses abundant, inexpensive LF data with scarce, costly HF data to achieve HF accuracy \cite{motamed2020multi,peherstorfer2018survey,zeng2024vehicle}. For instance, Feng et al. \cite{feng2024deep} proposed an ANN-based multi-fidelity deep learning framework to improve the efficiency of seismic structural analysis. Stavropoulou et al. \cite{stavropoulou2025multi} developed an LSTM-based multi-fidelity surrogate modeling approach for predicting nonlinear responses of multiple wave energy converters. Yang et al. \cite{yang2025data} introduced a multi-fidelity DeepONet framework capable of efficiently predicting spatio-temporal flow fields with significantly fewer high-fidelity data. Similarly, Zhong et al. \cite{zhong2025multi} presented a hybrid deep learning-enabled multi-fidelity framework to enhance time-series prediction in structural dynamic analysis. Zeng et al. \cite{zeng2023machine} also proposed a multi-fidelity temporal convolutional network (TCN) model for response prediction of vehicle crash tests. Although these methods demonstrate notable improvements in predictive accuracy and computational efficiency when compared to baseline models trained solely on LF data, they remain fundamentally deterministic. As a result, they do not provide any quantification of predictive uncertainty, which is critical for risk-informed decision-making and reliability assessment in high-stakes engineering applications.

To address this gap, a separate line of research has focused on probabilistic multi-fidelity modeling. However, existing approaches present their own set of challenges. For example, He et al. \cite{he2024active} proposed a multi-fidelity residual Gaussian process (GP) framework to probabilistically identify geomaterial properties. However, GP models are sensitive to kernel selection and hyperparameter estimation. Wang et al. \cite{wang2025multi} introduced a physics-guided generative adversarial network (GAN) for fatigue life prediction, where uncertainty in fatigue performance was also quantified. Nevertheless, GANs are notoriously difficult to train, often suffering from instability and mode collapse. More recently, Xie et al. \cite{xie2025multi} proposed a multi-fidelity Bayesian neural network (BNN) framework for aerodynamic modeling, where Hamiltonian Monte Carlo (HMC) was employed to estimate the posteriors of BNN hyperparameters. However, HMC is computationally expensive and scales poorly with the dimensionality of the parameter space. These limitations highlight a clear need for a multi-fidelity surrogate framework that is both probabilistic and computationally tractable, capable of providing robust uncertainty quantification without the training instabilities of GAN or the high computational overhead of traditional sampling-based approximate Bayesian inference methods.

Recent advances in generative artificial intelligence (AI) offer a promising new direction for probabilistic surrogate modeling. Generative AI models have demonstrated a remarkable capacity for learning high-dimensional probability distributions, capturing complex nonlinear relationships, and generating realistic synthetic data across domains like computer vision \cite{wang2021generative}, natural language processing \cite{iorliam2024comparative}, and scientific machine learning \cite{zeng2025solving}. These capabilities are uniquely suited to address the key challenges in probabilistic multi-fidelity modeling. First, as inherently probabilistic models, generative methods naturally quantify predictive uncertainty. Second, they can flexibly represent complex, nonlinear relationships between LF and HF data. Third, through deep neural architectures, generative AI is well-suited to handle high-dimensional input-output spaces (e.g., spatio-temporal fields, time-series responses), where probabilistic surrogates such as sampling-based BNN models often become computationally intractable. These properties make generative models a compelling foundation for developing robust probabilistic multi-fidelity surrogate frameworks.
Nevertheless, despite this clear potential, the application of generative modeling to multi-fidelity surrogate modeling remains limited. To the best of the authors’ knowledge, existing studies have primarily focused on traditional deep learning frameworks, with relatively few efforts dedicated to leveraging the power of generative AI.

To address this gap, this study proposes a generative AI–enhanced probabilistic multi-fidelity surrogate modeling framework. Our approach is designed in two stages: first, a generative surrogate is pretrained on abundant LF data to learn a base probabilistic model; second, theis model is fine-tuned using limited HF data via transfer learning, enabling it to capture complex LF--HF relationships. The proposed framework employs surjective normalizing flows (NF), a dimension-reducing variant of flow-based generative models. Unlike conventional NF, which are constrained to preserve input dimensionality in the latent space, surjective NF integrates both surjective and bijective layers within a unified neural architecture. The surjective layers project high-dimensional responses onto a lower-dimensional latent space, while the bijective layers ensure invertibility where needed, allowing exact density evaluation and tractable likelihood estimation. This hybrid property enables surjective NF to effectively handle high-dimensional responses such as time-series data, while providing the ability to train and evaluate the model on time-series data.
To the best of the authors' knowledge, the use of generative models for probabilistic multi-fidelity surrogate modeling has received limited attention in the existing literature. In contrast to most existing deep learning-based multi-fidelity approaches, which are predominantly deterministic, the proposed framework is probabilistic by construction, allowing for direct, principled quantification of predictive uncertainty associated with the use of multi-fidelity data.

The proposed multi-fidelity framework is applied to surrogate modeling of the time-series response of a rail-sleeper-ballast system and a reinforced concrete slab structure. The LF data are generated from a coarse-mesh FE model, whereas the HF data are obtained from a fine-mesh model. To evaluate the robustness of the framework, different levels of correlation between LF and HF data, as well as the effect of varying the amount of HF data, are systematically investigated. The results demonstrate that the proposed approach achieves efficient, HF-accurate probabilistic response predictions.

The remainder of this paper is organized as follows: in \cref{sec:methodology} we formulate the data-driven multi-fidelity surrogate modeling problem, and describe the proposed generative transfer learning approach we take to solve this problem.
In \cref{sec:NF} we describe the NF architecture for the probabilistic surrogate modeling, and in \cref{sec:application} we apply the proposed framework to two engineering benchmarks and analyze the performance of the framework under different levels of LF--HF data correlation and the effect of varying the number of HF training samples. In addition, we also benchmark the method against a TCN baseline \cite{zhong2025multi, zeng2023machine}. Finally, we summarize our conclusions and outline future research directions in \cref{sec:conclusion}.

\section{Probabilistic multi-fidelity surrogate modeling framework}
\label{sec:methodology}

\subsection{Problem formulation}
\label{sec:prob}

We restrict our attention in this work to the bi-fidelity setting.
Let \( m, n \in \mathbb{N} \), and let \(\Theta \coloneqq \mathbb{R}^m\) and \( Y \coloneqq \mathbb{R}^n \) denote the space of model parameters and the space of model outputs, respectively.
In this work we consider the problem of learning a surrogate model of the functional relation \(G \colon \Theta \to Y\) between model parameters and model outputs from data.
To generate the training data, we assume there is a data-generating distribution of model parameters \( p(\bm{\theta}) \) and that we have access to two data-generating models: an LF model \( G_{\mathrm{LF}} \colon \Theta \to Y \), and a HF model \( G_{\mathrm{HF}} \colon \Theta \to Y \).
Here we assume for simplicity that both the LF and HF outputs are in \( Y \).
These models define the data-generating distributions \( p_{\mathrm{LF}}(\mathbf{y}, \bm{\theta}) \) and \( p_{\mathrm{HF}}(\mathbf{y}, \bm{\theta}) \), respectively.

Let $\|\cdot\|_2$ denote the Euclidean norm.
Furthermore, we assume that the expected prediction error of the HF model with respect to the true model \(G\), \(\mathbb{E}_{p(\bm{\theta})} \left\| G_{\mathrm{HF}}(\bm{\theta}) - G(\bm{\theta}) \right\|_2\), is negligible, that the LF model is significantly biased, and that it is much computationally faster to sample from the LF data-generating distribution than from the HF data-generating distribution.
Therefore, our goal is to learn a model for \( p_{\mathrm{HF}}(\mathbf{y} \mid \bm{\theta}) \) by using both an LF dataset \( \mathcal{D}_{\mathrm{LF}} \coloneqq \{ \mathbf{y}^{(i)}_{\mathrm{LF}}, \bm{\theta}^{(i)} \}^{N_{\mathrm{LF}}}_{i = 1} \sim p_{\mathrm{LF}} (\mathbf{y}, \bm{\theta}) \) and a HF dataset \( \mathcal{D}_{\mathrm{HF}} \coloneqq \{ \mathbf{y}^{(i)}_{\mathrm{HF}}, \bm{\theta}^{(i)} \}^{N_{\mathrm{HF}}}_{i = 1} \sim p_{\mathrm{HF}} (\mathbf{y}, \bm{\theta}) \), with \(N_{\mathrm{LF}} \gg N_{\mathrm{HF}} \).

\subsection{Multi-fidelity generative modeling via transfer learning}
\label{sec:method}

\begin{figure}[htbp]
  \centering
  \includegraphics[width=\textwidth]{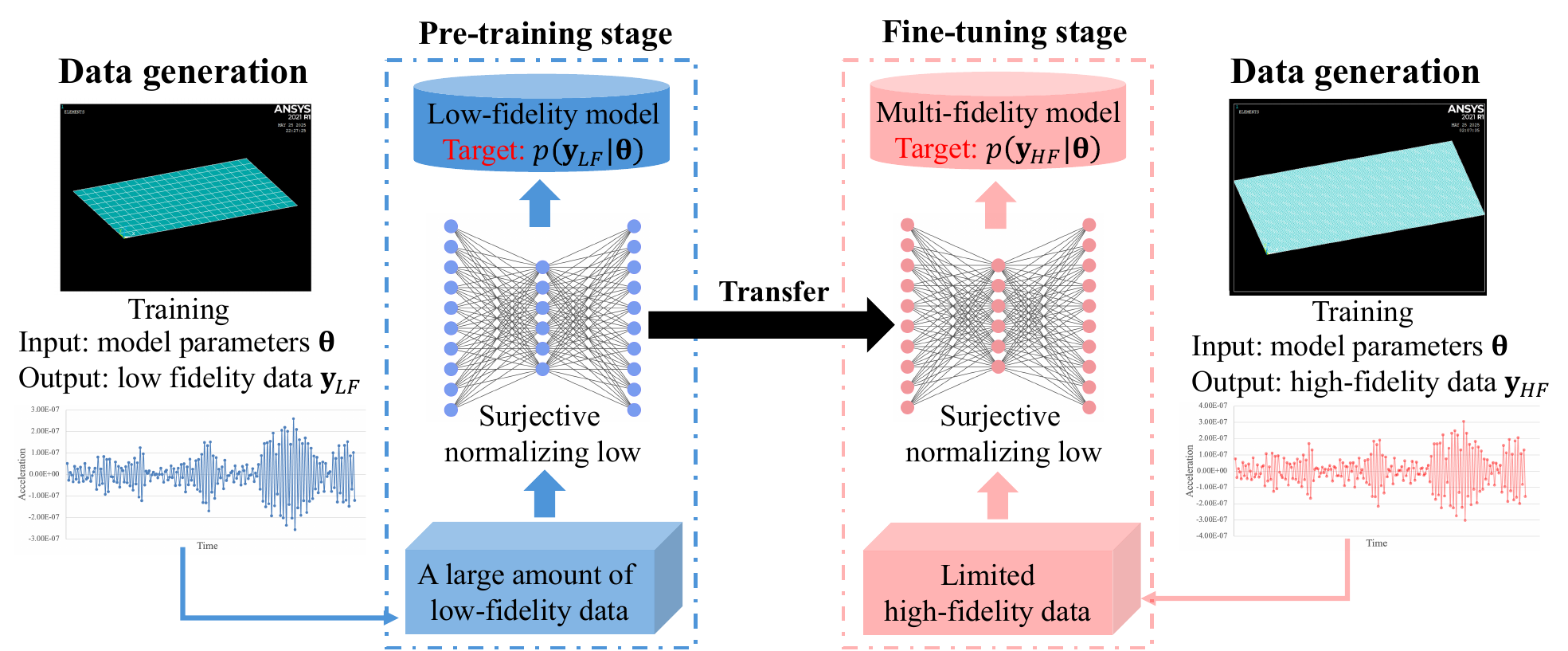}
  \caption{Workflow of the proposed probabilistic multi-fidelity surrogate modeling framework.}
  \label{fig:workflow}
\end{figure}

To solve this probabilistic multi-fidelity surrogate modeling problem we propose the framework illustrated in \Cref{fig:workflow}.
This framework consists of two stages: pre-training and fine-tuning. Both stages employ surjective NF models as the backbone, which function as probabilistic forward surrogates by approximating the conditional distributions $p_{\mathrm{LF}}(\mathbf{y} \mid \bm{\theta})$ and $p_{\mathrm{HF}}(\mathbf{y} \mid \bm{\theta})$.

In the pre-training stage, the LF dataset is used to train a surjective NF to approximate $p_{\mathrm{LF}} (\mathbf{y} \mid \bm{\theta})$.
This stage establishes a base generative model that captures the general behavior and response distributions of the low-fidelity system.
In the fine-tuning stage, we utilize the smaller HF dataset to fine-tune the pre-trained LF model and correct for the biases inherent in the LF data, resulting in a model approximating $p_{\mathrm{HF}}(\mathbf{y} \mid \bm{\theta})$.

A key advantage of this generative approach is that the resulting surrogate provides a full probabilistic distribution for any given input model parameters, rather than a single-point estimate. The proposed framework thus effectively addresses data scarcity by fusing abundant LF data with limited HF data. This process yields accurate, uncertainty-aware surrogates that captures HF-level physics while significantly reducing the computational burden associated with extensive HF simulations.

The use of transfer learning is motivated by the assumption that LF and HF models share a common underlying structure and differ due to model-form error or missing fine-scale dynamics.
Under this assumption, pre-training on abundant LF data allows the model to learn a meaningful representation of the mapping, which can then be adapted to the HF regime using a limited number of samples.
This strategy is particularly effective when the LF model provides a biased but informative approximation to the HF model.
However, it must be acknowledged that the effectiveness of transfer learning depends on the degree of correlation between LF and HF data.
If the LF model is poorly correlated with the HF model, or if the discrepancy between them is highly nonlinear or dominated by noise, the benefit of the proposed pre-training approach may be limited.
In such cases, fine-tuning may require more HF data, or alternative strategies may be needed.

Let \( q (\mathbf{y} \mid \bm{\theta}, \bm{\phi}) \) denote the parameterized surjective NF model, where \(\bm{\phi}\) denotes the model parameters.
In the pre-training stage, we learn the LF model by maximizing the expectation of the surjective NF model's likelihood over the LF data, that is,
\begin{equation*}
  \bm{\phi}_{\mathrm{LF}}^{\star} = \argmax_{\bm{\phi}_{\mathrm{LF}}} \ \mathbb{E}_{(\mathbf{y}, \bm{\theta}) \sim \mathcal{D}_{\mathrm{LF}}} \log q (\mathbf{y} \mid \bm{\theta}, \bm{\phi}_{\mathrm{LF}} ).
\end{equation*}
This step extracts the dominant structural patterns that are already well captured at the LF level.
Next, in the fine-tuning stage we transfer knowledge from the LF model to the MF model by using the LF-trained parameters $\bm{\phi}_{\mathrm{LF}}^{\star}$ as the initial guess for the parameters of the MF model, and then fine-tuning the MF model by maximizing its likelihood over the HF data, that is,
\begin{equation*}
  \bm{\phi}_{\mathrm{MF}}^{\star} = \argmax_{\bm{\phi}_{\mathrm{MF}}} \ \mathbb{E}_{(\mathbf{y}, \bm{\theta}) \sim \mathcal{D}_{\mathrm{HF}}} \log q_{\bm{\phi}} (\mathbf{y} \mid \bm{\theta}, \bm{\phi}_{\mathrm{MF}}, \bm{\phi}_{\mathrm{LF}}^{\star}).
\end{equation*}
where $\bm{\phi}_{\mathrm{MF}}^{\star}$ denotes the optimal parameters of the final MF model, and the notation \( q_{\bm{\phi}} (\mathbf{y} \mid \bm{\theta}, \cdot, \bm{\phi}_{\mathrm{LF}}^{\star}) \) indicates that the model parameters' optimization initial guess is taken to be  $\bm{\phi}_{\mathrm{LF}}^{\star}$.

Once trained, the MF surrogate can be queried for any new input $\bm{\theta}^{\ast}$.
Specifically, we use Monte Carlo sampling to generate samples from the predictive distribution \( q (\mathbf{y} \mid \bm{\theta}^{\ast}, \bm{\phi}_{\mathrm{MF}}^{\star}) \), which are then employed to estimate the predictive mean and variance, credibility intervals (CIs), and other statistics of interest.
This enables the surrogate not only to generate accurate HF predictions but also to quantify predictive uncertainty.

\section{NF-based probabilistic surrogate models}
\label{sec:NF}

The core of our proposed framework is the surjective conditional NF model employed to approximate conditional distributions of the form \( p(\mathbf{y} \mid \bm{\theta}) \). Surjective NFs extend the traditional NF architecture by integrating dimension-reducing (surjective) layers together with standard bijective transformations. This hybrid design is key to enabling scalable modeling of high-dimensional responses such as time-series data, a task that is often intractable for standard NF. To provide the necessary background, we first review bijective conditional NFs before describing the surjective architecture we employ in this work.

\subsection{Bijecive conditional NFs}
\label{sec:bi-NF}

Conditional NFs construct the conditional distribution $p(\mathbf{y} \mid \bm{\theta})$ by transporting a simple base density $p_{\mathbf{z}}(\mathbf{z})$ (with \( \mathbf{z} \in \mathbb{R}^n \), typically $\mathcal{N}(\mathbf{0},\mathbf{I}_n)$) through a parameterized invertible, $\bm{\theta}$-dependent map \cite{papamakarios2021normalizing}, that is,
\begin{equation}
  \label{eq:cnf-map}
  \mathbf{z} = f(\mathbf{y};\,\bm{\theta}, \bm{\phi}), \quad \mathbf{y} = f^{-1}(\mathbf{z};\,\bm{\theta}, \bm{\phi}), \quad \mathbf{z} \sim p_{\mathbf{z}}(\mathbf{z}),
\end{equation}
where $f(\cdot\,;\bm{\theta}, \bm{\phi})$ denotes the parameterized \(\mathbf{y} \to \mathbf{z}\) invertible transformation, $f^{-1}(\cdot\,;\bm{\theta}, \bm{\phi})$ its inverse, and $\bm{\phi}$ its parameters.
In practice, $f$ is constructed by composing $K$ invertible bijections \( \{f_k\}_{k=1}^K \), that is, we employ the sequence of transformations
\begin{equation*}
  \mathbf{y} \coloneqq \mathbf{z}_0 \to \mathbf{z}_1 \to \dots \to \mathbf{z}_{K - 1} \to \mathbf{z}_K \coloneqq \mathbf{z},
\end{equation*}
where the \(\mathbf{z}_k\)s are given by
\begin{equation}
  \label{eq:cnf-composition}
  \mathbf{z}_k \;=\; f_k\!\big(\mathbf{z}_{k-1};\,\bm{\theta}, \bm{\phi}\big), \quad  k=1,\dots,K.
\end{equation}
Applying the multivariate change-of-variables formula to \cref{eq:cnf-map,eq:cnf-composition} leads to the closed-form expression for \( p(\mathbf{y} \mid \bm{\theta}) \)
\begin{equation}
  \label{eq:cnf-density}
  p(\mathbf{y} \mid \bm{\theta}, \bm{\phi}) = p_{\mathbf{z}}(\mathbf{z}_K)\; \prod_{k=1}^{K} \left|\det \frac{\partial f_k}{\partial \mathbf{z}_{k-1}} \left (\mathbf{z}_{k-1};\,\bm{\theta}, \bm{\phi} \right) \right|,
\end{equation}
or, in log form,
\begin{equation}
  \label{eq:cnf-loglik}
  \log p(\mathbf{y} \mid \bm{\theta}, \bm{\phi}) = \log p_{\mathbf{z}}(\mathbf{z}_K) + \sum_{k=1}^{K} \log\left|\det \frac{\partial f_k}{\partial \mathbf{z}_{k-1}} \left( \mathbf{z}_{k-1};\,\bm{\theta} \right) \right|.
\end{equation}
The components of the flow transformation are chosen so that each log-determinant term is inexpensive to evaluate (e.g., via triangular Jacobians), yielding an explicit, tractable closed-form expression for the log-likelihood as a function of the flow parameters \(\bm{\phi}\).

Given training data $\mathcal{D} \coloneqq \{(\mathbf{y}^{(i)}, \bm{\theta}^{(i)})\}_{i=1}^{N}$, the flow parameters $\bm{\phi}$ are estimated by maximizing the average log-likelihood over the training data, that is,
\begin{equation}
\label{eq:cnf-obj}
\begin{aligned}
\mathcal{L}(\bm{\phi})
&= \frac{1}{N}\sum_{i=1}^{N} \log p\!\left(\mathbf{y}^{(i)} \mid \bm{\theta}^{(i)}, \bm{\phi}\right) \\
&= \frac{1}{N}\sum_{i=1}^{N}
\left[
\log p_{\mathbf{z}}\!\big(f(\mathbf{y}^{(i)}; \bm{\theta}^{(i)}, \bm{\phi})\big)
+ \sum_{k=1}^{K}
\log \left| \det \frac{\partial f_k(\mathbf{z}_{k-1}; \bm{\theta}^{(i)}, \bm{\phi})}{\partial \mathbf{z}_{k-1}} \right|
\right],
\end{aligned}
\end{equation}
which is optimized using stochastic gradient methods.

To generate predictions for a new input $\bm{\theta}^\ast$, we employ the sampling procedure implied by \cref{eq:cnf-map}, that is, to draw samples from the distribution of latent codes, $\mathbf{z} \sim p_{\mathbf{z}}$, and apply the inverse transformation $f^{-1}(\mathbf{z};\bm{\theta}^\ast, \bm{\phi})$, to these samples, resulting in an ensemble of draws from the predictive distribution $p(\mathbf{y} \mid \bm{\theta}^\ast, \bm{\phi})$.
These generated samples can then be used to compute statistics of the predictive distribution such as mean, variance, and predictive means, among others, and thus provide a complete probabilistic forecast.

A critical limitation of the NF described above is their dimension-preserving property. Defining the flow in terms of a sequence of bijective transformations  mandates that the latent space $\mathbf{z}$ must have the same dimension as the data space $\mathbf{y}$.
This property becomes computationally prohibitive and inefficient when modeling high-dimensional structural responses such as time-series data.
To address this bottleneck, we propose employing surjective NFs to construct the proposed multifidelity probabilistic surrogates. This architecture explicitly augments the standard bijective stack with dimension-reducing (surjective) maps, enabling the model to efficiently compress the high-dimensional response into a low-dimensional latent representation, thereby enhancing scalability for modeling high-dimensional responses.

\subsection{Surjective conditional NFs}
\label{sec:surjective-NF}

Surjective NFs extend the classical bijective flow framework by allowing for dimension-reducing transformations, thereby connecting NFs with variational autoencoders. This flexibility enables scalable likelihood-based training for high-dimensional responses, for which strictly bijective transformations may be computationally inefficient or overly restrictive. To provide the necessary methodological background, we briefly summarize the formulation adopted in this study based on the SurVAE framework of Nielsen et al. (2020) \cite{nielsen2020survae}. SurVAE provides a unified probabilistic framework for constructing generative models from bijective and surjective transformations. Within this framework, the data log-likelihood is decomposed into contributions involving the base density, Jacobian corrections for bijective transformations, and probabilistic encoding or decoding terms for surjective transformations.

Let again \( \mathbf{y} \in \mathbb{R}^W \) denote an output datum  and introduce a latent variable \( \mathbf{z}\in\mathbb{R}^{Q}\), \( Q\le W\), with a base distribution \( p_{\mathbf{z}}(\mathbf{z}) \).
Furthermore, we relate \( \mathbf{z} \) and \( \mathbf{y} \) via the surjective map  \( h \colon \mathbb{R}^W \to \mathbb{R}^Q \), \( \mathbf{z} = h(\mathbf{y}; \bm{\theta}) \), parameterized by the input data \( \bm{\theta} \) (contrast with \cref{eq:cnf-map}, where the \( \mathbf{y} \to \mathbf{z} \) is bijective.)
Given that the \( \mathbf{y} \to \mathbf{z} \) is not bijective, we cannot employ the change-of-variables formula to derive a closed-form expression for \(\log p(\mathbf{y} \mid \bm{\theta} )\) as in \cref{eq:cnf-loglik}.
Instead, in the SurVAE framework, we consider the inverse of the surjection \( h \) to be a stochastic transformation \( p(\mathbf{y} \mid \mathbf{z}, \bm{\theta}) \), and the log-likelihood of the output data is modeled in terms of the stochastic and surjective maps as \cite{nielsen2020survae,dirmeiersimulation}
\begin{equation}
  \label{eq:survae-decomp}
  \log p(\mathbf{y} \mid \bm{\theta}) = \log p_{\mathbf{z}}(\mathbf{z}) + V(\mathbf{y},\mathbf{z}  \mid \bm{\theta}) + E(\mathbf{y},\mathbf{z} \mid \bm{\theta}), \quad \mathbf{z}\sim q(\mathbf{z} \mid \mathbf{y}, \bm{\theta}),
\end{equation}
where \( q(\mathbf{z} \mid \mathbf{y}, \bm{\theta}) \) is a variational distribution, \( E(\mathbf{y}, \mathbf{z}) \) denotes the variational bound looseness, and \( V(\mathbf{y}, \mathbf{z}) \) denotes the ``likelihood contribution'', given by
\begin{equation}
\label{eq:V-limit}
V(\mathbf{y},\mathbf{z} \mid \bm{\theta}) = \lim_{q(\mathbf{z} \mid \mathbf{y},\bm{\theta}) \to \delta(\mathbf{z} - h(\mathbf{y}; \bm{\theta}))} \mathbb{E}_{q(\mathbf{z} \mid \mathbf{y})} \left[ \log \frac{p(\mathbf{y} \mid \mathbf{z},\bm{\theta})} {q(\mathbf{z} \mid \mathbf{y}, \bm{\theta} )} \right ].
\end{equation}
The bound looseness term vanishes if the so-called ``right inverse condition'' is satisfied \cite{klein2021funnels, nielsen2020survae}, that is, if \( p(\mathbf{y} \mid \mathbf{z}, \bm{\theta}) \) defines a distribution over the possible right inverses of the surjection \(h\).

It remains to formulate the surjective layer architecture by constructing the surjection \( \mathbf{y} \to \mathbf{z} \) and the stochastic transformation \( \mathbf{z} \to \mathbf{y} \).
To do so, we split $\mathbf{y}$ into discarded and kept blocks:
\begin{equation}
\label{eq:split}
\mathbf{y} = \begin{bmatrix}\mathbf{y}^{-}\\\mathbf{y}^{+}\end{bmatrix}, \quad \mathbf{y}^{-}\in\mathbb{R}^{W-Q}, \quad \mathbf{y}^{+}\in\mathbb{R}^{Q}, \quad 0<Q<W.
\end{equation}
and define a conditional bijection on the kept block
\begin{equation}
\label{eq:f-def}
f \colon \mathbb{R}^{Q} \to \mathbb{R}^{Q}, \quad \mathbf{y}^{+} = f(\mathbf{z} ;\mathbf{y}^{-},\bm{\theta}), \quad \mathbf{z} = f^{-1}(\mathbf{y}^{+};\mathbf{y}^{-},\bm{\theta}),
\end{equation}
where $f$ is any tractable flow transform (e.g., coupling or autoregressive) conditional on $(\mathbf{y}^{-},\bm{\theta})$.
Intuitively, the pair $(\mathbf{y}^{-},\bm{\theta})$ selects a bijection acting on $\mathbf{y}^{+}$.
Therefore, in the \( \mathbf{y} \to \mathbf{z} \), the surjective layer splits \( \mathbf{y} \) into blocks as in \cref{eq:split} and deterministically generates \( \mathbf{z} \) from the blocks using the flow transformation \( f \) as in \cref{eq:f-def}.
In the \( \mathbf{z} \to \mathbf{y} \) direction, a surjective layer then: (i) stochastically generates \( \mathbf{y}^{-} \) by sampling from the stochastic ``decoder'' \( p(\mathbf{y}^{-} \mid \mathbf{z},\bm{\theta}) \), and (ii) deterministically generates \( \mathbf{y}^{+} \) using the flow transformation \( f \) as in \cref{eq:f-def}.

Based on this construction, it can be said that the conditional likelihood factorizes the distribution of the outputs into a stochastic decoder on the discarded block and a Dirac delta distribution on the kept block, that is,
\begin{equation}
\label{eq:layer-factorization-ov}
p(\mathbf{y} \mid \mathbf{z},\bm{\theta}) = p(\mathbf{y}^{-} \mid \mathbf{z},\bm{\theta}) \, \delta\!\left ( \mathbf{y}^{+}-f(\mathbf{z};\mathbf{y}^{-},\bm{\theta}) \right ).
\end{equation}
Furthermore, limiting distribution for the variational density is given by the Dirac delta change-of-variables rule as
\begin{equation}
  \label{eq:q-delta-ov}
  \begin{split}
    q(\mathbf{z} \mid \mathbf{y}) &\to \delta\!\left ( \mathbf{z}-f^{-1}(\mathbf{y}^{+};\mathbf{y}^{-},\bm{\theta}) \right ) \\
    &= \delta\!\left (\mathbf{y}^{+}-f(\mathbf{z};\mathbf{y}^{-},\bm{\theta}) \right ) \left | \det J^{-1} \right |^{-1},
  \end{split}
\end{equation}
where \( J(\mathbf{z}; \mathbf{y}^{-}, \bm{\theta}) = \partial f(\mathbf{z};\mathbf{y}^{-},\bm{\theta}) / \partial \mathbf{z} \) is the Jacobian of the flow transformation.
Substituting \cref{eq:layer-factorization-ov}--\cref{eq:q-delta-ov} into \cref{eq:V-limit}, the likelihood contribution is expressed as \cite{klein2021funnels}
\begin{equation}
  \label{eq:V-eval-steps}
  \begin{split}
    V(\mathbf{y},\mathbf{z} \mid \bm{\theta})
    & = \int  \mathrm{d} \mathbf{z} \delta\! \left ( \mathbf{z} - f^{-1}(\mathbf{y}^{+};\mathbf{y}^{-},\bm{\theta}) \right ) \log \left [ \frac{ p(\mathbf{y}^{-} \mid \mathbf{z},\bm{\theta}) \, \delta\!\left ( \mathbf{y}^{+}-f(\mathbf{z};\mathbf{y}^{-},\bm{\theta}) \right ) }{ \delta\!\left ( \mathbf{y}^{+}-f(\mathbf{z};\mathbf{y}^{-},\bm{\theta}) \right )\, \left |\det J^{-1}\right |^{-1} } \right ] \\
    &= \log p(\mathbf{y}^{-} \mid \mathbf{z}, \bm{\theta}) - \log \left |\det J(\mathbf{z};\mathbf{y}^{-}, \bm{\theta})\right |, \quad \text{s.t.} \quad \mathbf{y}^{+}=f(\mathbf{z};\mathbf{y}^{-},\bm{\theta}).
  \end{split}
\end{equation}
For this construction, the looseness term vanishes \cite{klein2021funnels}.
Therefore, \Cref{eq:survae-decomp} becomes
\begin{equation}
\label{eq:single-layer-lik-ov}
\begin{aligned}
\log p(\mathbf{y} \mid \bm{\theta})
&= \log p_{\mathbf{z}} (\mathbf{z}) + V(\mathbf{y},\mathbf{z}) \\[4pt]
&= \underbrace{\log p_{\mathbf{z}}(\mathbf{z})}_{\text{base}}
   \;+\;
   \underbrace{\log p(\mathbf{y}^{-} \mid \mathbf{z},\bm{\theta})}_{\text{decoder on discarded}}
   \;-\;
   \underbrace{\log \left |\det J(\mathbf{z};\mathbf{y}^{-},\bm{\theta})\right |}_{\text{kept-block Jacobian}},\\
&\quad \text{s.t.} \quad \mathbf{y}^{+} = f(\mathbf{z};\mathbf{y}^{-},\bm{\theta}).
\end{aligned}
\end{equation}
This formula is then used to compute the single-layer likelihood during training and inference.
In this work, we employ the decoder architecture
\begin{equation}
  \label{eq:decoder-mlp}
  p(\mathbf{y}^{-} \mid \mathbf{z}, \bm{\theta}, {\bm{\phi}})
  \coloneqq
  \mathcal{N}\!\left(
  \mu(\mathbf{z}; \bm{\theta}, {\bm{\phi}}),
  \operatorname{diag}\!\left(\exp\!\big(2\,\sigma(\mathbf{z}; \bm{\theta}, {\bm{\phi}})\big)\right)
  \right),
\end{equation}
where $\mu$ and $\sigma$ denote the outputs of a multilayer perceptron parameterized by $\bm{\phi}$.

We now proceed to integrate the surjective layers defined above with bijective layers.
For evaluating the likelihood of a given output \( \mathbf{y} \), we evaluate a $K$-layer flow consisting of both bijective (dimension-preserving) and surjective (dimension-reducing) layers. We work backwards through these layers to the base latent variable \(\mathbf{u}_{0} \sim p_{0} \).
To avoid overloading the symbol \(\mathbf{z}\), which is reserved for the lower-dimensional variables associated with the surjective layers, we denote the intermediate flow states by \(\mathbf{y} \equiv \mathbf{u}_{K}, \mathbf{u}_{K-1}, \dots, \mathbf{u}_{1}, \mathbf{u}_{0}\).
We partition the layer indices into bijective and surjective layer indices as
\begin{equation}
  \label{eq:layer-sets}
  \mathcal{B}=\{\text{bij. layers}\}, \ \mathcal{S}=\{\text{surj. layers}\}, \ \mathcal{B}\cap\mathcal{S}=\varnothing, \ \mathcal{B}\cup\mathcal{S}=\{1,\dots,K\}.
\end{equation}

Bijective layers employ the parameterized bijection \( \mathbf{u}_{k-1}=f_{k}^{-1}(\mathbf{u}_{k};\bm{\theta}, \bm{\phi}) \) and contribute only the standard change-of-variables Jacobian. Let \( J_{k}^{(\mathrm{bij})} = \partial f_{k}(\mathbf{u}_{k-1};\bm{\theta}, \bm{\phi}) / \partial \mathbf{u}_{k-1} \).
This contribution takes the form
\begin{equation}
\label{eq:bijective-contrib}
C_{k}^{(\mathrm{bij})}
=-\log \left | \det J_{k}^{(\mathrm{bij})} \right |_{\mathbf{u}_{k-1}=f_{k}^{-1}(\mathbf{u}_{k};\bm{\theta}, \bm{\phi})}.
\end{equation}

A surjection splits the state into kept and discarded blocks \( \mathbf{u}^{+}_k \) and \( \mathbf{u}^{-}_k \), respectively, and employs the parameterized kept-block bijection \( f_{k}(\cdot\,;\mathbf{u}_{k}^{-},\bm{\theta}, \bm{\phi}) \) and the decoder \( p_{k} \left(\cdot \mid \mathbf{z}_{k},\bm{\theta}, \bm{\phi} \right) \). Define
\begin{equation}
  \label{eq:surj-def}
  \mathbf{z}_{k}
  =f_{k}^{-1}(\mathbf{u}_{k}^{+};\mathbf{u}_{k}^{-},\bm{\theta}, \bm{\phi}),
  \quad
  J_{k}^{(\mathrm{surj})}
  =\partial f_{k}(\mathbf{z}_{k};\mathbf{u}_{k}^{-},\bm{\theta}, \bm{\phi}) / \partial \mathbf{z}_{k}.
\end{equation}

From \cref{eq:single-layer-lik-ov}, surjective layers contribute both a decoder log-density for the discarded block and a kept-block Jacobian term.
These contributions take the form
\begin{equation}
\label{eq:surj-contrib}
C_{k}^{(\mathrm{surj})}
=\log p_{k} \left(\mathbf{u}_{k}^{-} \mid \mathbf{z}_{k},\bm{\theta}, \bm{\phi} \right)
-\log \left |\det J_{k}^{(\mathrm{surj})} \right |.
\end{equation}
After the surjective transformation, the next-layer state is given by \(\mathbf{u}_{k-1} \coloneqq \mathbf{z}_{k}\).

Collecting the contributions from all layers $k=K,\dots,1$ yields the log-conditional likelihood of the output \( \mathbf{y} \) given an input \( \mathbf{\theta} \):
\begin{equation}
\label{eq:full-lik}
\log q(\mathbf{y} \mid \bm{\theta}, \bm{\phi})
=\underbrace{\log p_{0}(\mathbf{u}_{0})}_{\text{base density}}
+\sum_{k\in\mathcal{B}}
\underbrace{C_{k}^{(\mathrm{bij})}}_{\text{Eq.\,\eqref{eq:bijective-contrib}}}
+\sum_{k\in\mathcal{S}}
\underbrace{C_{k}^{(\mathrm{surj})}}_{\text{Eq.\,\eqref{eq:surj-contrib}}}.
\end{equation}
This expression provides provides a tractable expression for every training example \( (\mathbf{y}, \bm{\phi}) \).

The parameters \( \bm{\phi} \) are estimated by maximizing the expected log-likelihood over a training dataset \( \mathcal{D} \), that is,
\begin{equation}
  \label{eq:mc-expectation}
  \mathcal{L}(\bm{\phi}) \coloneqq \mathbb{E}_{(\mathbf{y}, \bm{\theta}) \sim \mathcal{D}} \left [ \log q(\mathbf{y} \mid \bm{\theta}, \bm{\phi}) \right], \quad \bm{\phi}^{\ast} \coloneqq \argmax_{\bm{\phi}} \mathcal{L}(\bm{\phi}).
\end{equation}
Accordingly, the gradient of the objective is given by
\begin{equation}
  \label{eq:grad-estimator}
  \nabla_{\bm{\phi}}\mathcal{L}(\bm{\phi}) = \mathbb{E}_{(\mathbf{y}, \bm{\theta}) \sim \mathcal{D}} \, \nabla_{\bm{\phi}} \log q (\mathbf{y} \mid \bm{\theta}, \bm{\phi}),
\end{equation}
which is directly estimated via automatic differentiation through the flow architecture.

It is worth noting that conditional variational autoencoders (cVAEs) are also used for dimensionality reduction and uncertainty quantification \cite{mylonas2021conditional}. However, cVAEs learn the latent representation through variational approximations which may introduce approximation bias and posterior collapse \cite{doersch2016tutorial}. In contrast, the proposed surjective NF performs dimensionality reduction through explicit surjective transformations while retaining exact likelihood evaluation. Rather than relying on variational inference, it enables exact density evaluation and sampling, avoiding approximation bias and providing more accurate likelihood estimation. From this perspective, surjective NFs combines features of both NFs and VAEs, making them well suited for multi-fidelity surrogate modeling.

\subsection{Flow architecture}\label{sec:flow-architecture}

In this study, the flow architecture is composed of two types of transformations: a masked coupling transformation for dimension-preserving mappings \cite{dinh2016density,papamakarios2017masked} and a masked coupling inference funnel for dimension-reducing mappings \cite{klein2021funnels}.
implemented using the \textsc{surjectors} library \cite{dirmeier2024surjectors}.
Given a prescribed sequence of layer dimensions, a layer is implemented as a masked coupling transformation when the input and output dimensions are the same, and as a masked coupling inference funnel when the output dimension is smaller than the input dimension. Between successive layers, alternating binary masks are used so that the coordinates kept fixed in one layer are transformed in the next. The final latent variable follows a standard Gaussian base distribution.

For a dimension-preserving coupling layer, let
\(\mathbf{u}_{k-1}\in\mathbb{R}^{d_k}\) denote the input and let
\(\mathbf{m}_k\in\{0,1\}^{d_k}\) be a binary mask, where
\(m_{k,j}=1\) indicates that the \(j\)th coordinate remains unchanged.
The mask partitions the coordinate indices into an unchanged set \(A_k\)
and a transformed set \(B_k\). For a given vector
\(\mathbf{a}\in\mathbb{R}^{d_k}\), the subvectors containing the
coordinates indexed by \(A_k\) and \(B_k\) are denoted by
\(\mathbf{a}_{A_k}\) and \(\mathbf{a}_{B_k}\), respectively.
The unchanged block is used as the conditioning input to a multilayer
perceptron, which produces the elementwise log-scale and shift vectors,
\begin{equation}
\mathbf{s}_k
=\mathbf{s}_k\!\left(\mathbf{u}_{k-1,A_k}\right),
\qquad
\mathbf{t}_k
=
\mathbf{t}_k\!\left(\mathbf{u}_{k-1,A_k}\right).
\label{eq:mcf-conditioner}
\end{equation}
The masked affine coupling transformation is then given by
\begin{equation}
  \label{eq:mcf-forward}
\mathbf{u}_{k,A_k}
=\mathbf{u}_{k-1,A_k},
\qquad
\mathbf{u}_{k,B_k}
=\mathbf{u}_{k-1,B_k}
\odot
\exp\!\left(\mathbf{s}_k\right)
+
\mathbf{t}_k,
\end{equation}
and its inverse by
\begin{equation*}
\mathbf{u}_{k-1,A_k}
=
\mathbf{u}_{k,A_k},
\qquad
\mathbf{u}_{k-1,B_k}
=
\left(
\mathbf{u}_{k,B_k}
-
\mathbf{t}_k
\right)
\odot
\exp\!\left(-\mathbf{s}_k\right),
\end{equation*}
where \(\mathbf{s}_k\) and \(\mathbf{t}_k\) are evaluated using the
unchanged block
\(\mathbf{u}_{k,A_k}=\mathbf{u}_{k-1,A_k}\).

Because the unchanged block \(\mathbf{u}_{k,A_k}\) does not depend on
\(\mathbf{u}_{k-1,B_k}\), so the upper-right block is zero. Thus, only one off-diagonal block is zero, which makes the Jacobian has the block-triangular form
\begin{equation}
\frac{\partial \mathbf{u}_k}
{\partial \mathbf{u}_{k-1}}
=
\begin{bmatrix}
\mathbf{I} & \mathbf{0} \\[4pt]
\displaystyle
\frac{\partial \mathbf{u}_{k,B_k}}
{\partial \mathbf{u}_{k-1,A_k}}
&
\operatorname{diag}\!\left(\exp(\mathbf{s}_k)\right)
\end{bmatrix}.
\label{eq:mcf-jacobian}
\end{equation}
The lower-left block is generally nonzero because the scale and shift
vectors depend on the unchanged coordinates. However, this block does
not affect the determinant of the block-triangular Jacobian.
Consequently,
\begin{equation}
\log
\left|
\det
\frac{\partial \mathbf{u}_k}
{\partial \mathbf{u}_{k-1}}
\right|
=
\sum_{j\in B_k}s_{k,j}.
\label{eq:mcf-logdet}
\end{equation}
Therefore, the masked affine coupling transformation is bijective,
computationally inexpensive to invert, and permits exact likelihood
evaluation.

When the target dimension decreases, the architecture switches to a masked coupling inference funnel. This layer combines an internal masked coupling transformation with a conditional decoder so that dimensionality can be reduced while preserving exact likelihood computation. Let \( f_k \) denote the internal masked coupling map. In the inverse direction,
given the higher-dimensional state
\( \mathbf{u}_{k-1}\in\mathbb{R}^{d_{k-1}} \), we first compute
\begin{equation}
\tilde{\mathbf{u}}_{k-1}=f_k^{-1}(\mathbf{u}_{k-1};\bm{\theta},\bm{\phi}),
\label{eq:funnel-inverse}
\end{equation}
and then partition it as
\begin{equation}
\tilde{\mathbf{u}}_{k-1}
=\begin{bmatrix}
\mathbf{z}_k \\
\mathbf{u}_k^{-}
\end{bmatrix},
\quad
\mathbf{z}_k\in\mathbb{R}^{r_k},
\quad
\mathbf{u}_k^{-}\in\mathbb{R}^{d_{k-1}-r_k},
\label{eq:funnel-partition}
\end{equation}
where \( \mathbf{z}_k \) is the retained lower-dimensional variable and \( \mathbf{u}_k^{-} \) is the discarded block. The discarded block is modeled by a conditional decoder density \( p_k(\mathbf{u}_k^{-}\mid\mathbf{z}_k,\bm{\theta}, \bm{\phi})\).
By introducing the auxiliary variable \( \tilde{\mathbf{u}}_{k-1} \), the likelihood of \( \mathbf{u}_{k-1} \) follows from the chain rule and the change-of-variables formula. Hence, relative to the next layer, the contribution of the surjective layer to the log-likelihood is given by \Cref{eq:surj-contrib}

In the generative direction, the masked coupling inference funnel is
evaluated in reverse order. Given the reduced latent variable
\(\mathbf{z}_k\), the discarded block is sampled from the conditional diagonal Gaussian decoder
\begin{equation}
  \label{eq:conditional_decoder}
  p_k(\mathbf{u}_k^{-}\mid\mathbf{z}_k,\bm{\theta};\bm{\phi})
  =\mathcal{N}\!\left(
    \mathbf{u}_k^{-};
    \bm{\mu}_k(\mathbf{z}_k,\bm{\theta};\bm{\phi}),
    \operatorname{diag}\!\left \{ \left [
      2 \bm{\rho}_k(\mathbf{z}_k,\bm{\theta};\bm{\phi})
    \right] \right \}
  \right),
\end{equation}
where the vectors of conditional means \( \bm{\mu}_k(\mathbf{z}_k,\bm{\theta};\bm{\phi}) \) and log-standard deviations \(\bm{\rho}_k(\mathbf{z}_k,\bm{\theta};\bm{\phi})\) are given by the outputs of a two-headed multi-layer perceptron parameterized by \(\bm{\phi}\).
A sample of \(\mathbf{u}_k^{-}\) is then drawn from this distribution.
The retained and sampled blocks are then concatenated to form \(\tilde{\mathbf{u}}_{k-1} =[\mathbf{z}_k^\top,(\mathbf{u}_k^{-})^\top]^\top\), after which the forward masked coupling transformation \(f_k\) is applied to reconstruct the higher-dimensional state \(\mathbf{u}_{k-1}\). In this manner, the surjective mapping is directly integrated with the bijective masked coupling layer: the masked coupling transformation provides a tractable triangular Jacobian and exact change-of-variables correction, whereas the conditional decoder provides a probabilistic model for the coordinates removed during dimensionality reduction.

\section{Application}\label{sec:application}
This section evaluates the proposed probabilistic MF surrogate modeling framework for two structural dynamics applications: a rail--sleeper--ballast system and a reinforced-concrete slab. The two studies examine the ability of the proposed method to transfer information across fidelities while providing probabilistic response predictions and uncertainty estimates. The effects of the level of LF--HF discrepancy and the amount of available HF training data are also investigated to evaluate the robustness and data efficiency of the proposed framework.

\subsection{Example 1: Rail-sleeper-ballast system}\label{sec:rail}

\begin{figure}[htbp]
  \centering
  \includegraphics[width=\textwidth]{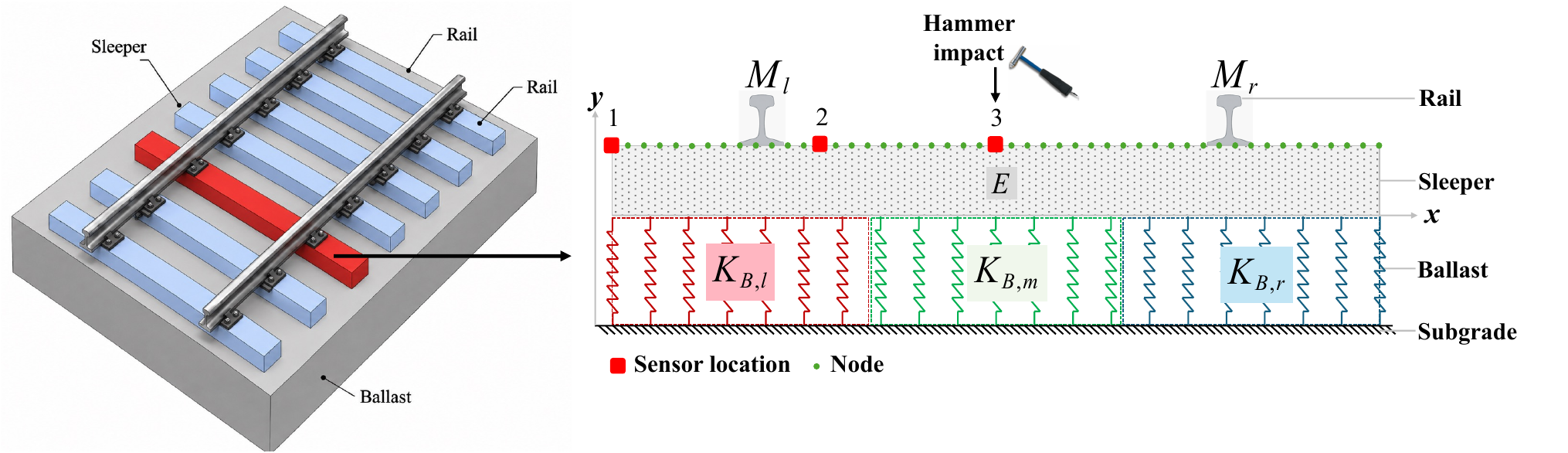}
  \caption{The rail--sleeper--ballast system. The sleeper is modeled using Timoshenko beam elements supported by three ballast regions with stiffnesses \(K_{B,l}\), \(K_{B,m}\), and \(K_{B,r}\), while the rails are represented by lumped masses at the rail-seat locations. The green markers denote the finite-element nodes through the prescribed spatial discretization of the sleeper, and the red markers indicate the selected sensor locations. A vertical hammer impact is applied at the midpoint of the sleeper.}
  \label{fig:rail_model}
\end{figure}

The first numerical example considers a rail--sleeper--ballast system, which is a representative structural component in railway infrastructure. In this system, the rails directly carry the external dynamic load and transfer it to the sleepers, while the sleepers further distribute the load to the ballast layer and supporting foundation.
As shown in \Cref{fig:rail_model}, the rail--sleeper--ballast system is modeled as a discretized dynamic system. The sleeper is represented using Timoshenko beam elements, while the ballast foundation is modeled as distributed vertical support stiffness along the beam. The ballast layer is divided into three regions, namely the left, middle, and right regions, with corresponding stiffness parameters \(k_{B,l}\), \(k_{B,m}\), and \(k_{B,r}\). The two rails are represented through lumped rail masses applied at the left and right rail-seat locations, denoted by \(M_l\) and \(M_r\), respectively. In addition, \(E\) denotes the elastic modulus of the sleeper. Therefore, the uncertain physical parameter vector is defined as
\begin{equation}
    \bm{\theta}
    =[k_{B,l}, k_{B,m}, k_{B,r}, E, M_l, M_r]^{\mathrm{T}} .
\end{equation}

The dynamic response is governed by the semi-discrete equation of motion
\begin{equation}
    \mathbf{M}(\bm{\theta})\ddot{\mathbf{u}}(t)
    + \mathbf{C}(\bm{\theta})\dot{\mathbf{u}}(t)
    + \mathbf{K}(\bm{\theta})\mathbf{u}(t)
    =\mathbf{f}(t),
\label{eq:motion_eq}
\end{equation}
where \(\mathbf{M}\), \(\mathbf{C}\), and \(\mathbf{K}\) are the assembled mass, damping, and stiffness matrices, respectively, \(\mathbf{u}(t)\) is the displacement vector, and \(\mathbf{f}(t)\) is the external force vector. 
The nominal ballast stiffness is \(150\times 10^6~\mathrm{N/m^2}\), and the nominal rail mass is \(42~\mathrm{kg}\). The parameters in \(\bm{\theta}\) are used as dimensionless scaling factors with respect to these nominal values.
For data generation, the six uncertain parameters are independently sampled from univariate uniform distributions \(\mathcal{U}(0.5,1.5)\). Thus, the generated samples account for variability in the three ballast stiffness regions, the sleeper elastic modulus, and the left and right rail masses.

Rayleigh damping is adopted in the dynamic simulation. The damping matrix is defined as \(\mathbf{C} = \alpha \mathbf{M} + \beta \mathbf{K}
\), where the Rayleigh coefficients \(\alpha\) and \(\beta\) are determined using the first and third vibration modes. A damping ratio of \(5\%\) is assigned to both selected modes.

The system is subjected to a vertical harmonic force excitation applied at the midpoint of the sleeper,
\begin{equation}
    f(t) = F_0 \sin(2\pi f_{\mathrm{exc}} t),
\end{equation}
where the force amplitude is \(F_0=3000~\mathrm{N}\) and the excitation frequency is \(f_{\mathrm{exc}}=180~\mathrm{Hz}\). The time-domain response is computed using the Newmark average-acceleration method \cite{craig2006fundamentals} with \(\beta=1/4\) and \(\gamma=1/2\).
The sampling frequency is \(f_s=6400~\mathrm{Hz}\), and the total simulation duration is \(T=0.03~\mathrm{s}\), resulting in \(n_t=192\) time steps for each response trajectory.

Three sensors are selected to record the acceleration responses. Specifically, the sensors are placed at the left end, the one-quarter location from the left end, and the midpoint of the sleeper, as shown in \Cref{fig:rail_model}. The model response is therefore a multivariate time-series signal. 192 time steps are recorded at each sensor, the response vector can be written as
\begin{equation}
    \mathbf{y}
    =\left[
    \mathbf{a}_1^{\mathrm{T}},
    \mathbf{a}_2^{\mathrm{T}},
    \mathbf{a}_3^{\mathrm{T}}
    \right]^{\mathrm{T}}
    \in \mathbb{R}^{3\times192},
\end{equation}
where \(\mathbf{a}_j\) denotes the acceleration time history measured by the \(j\)-th sensor. 

In this example, the fidelity gap is due to both the spatial discretization of the low- and high-fidelity models and the physical simplification of the rail--sleeper--ballast model. The HF model uses a finer 48-element Timoshenko beam discretization and preserves the spatially heterogeneous ballast stiffness and asymmetric rail mass distribution. Specifically, the HF response is generated as \( \mathbf{y}_{\mathrm{HF}} =\mathcal{M}_{48}(\bm{\theta}) \),
where \(\mathcal{M}_{48}(\cdot)\) denotes the 48-element dynamic model.

On the other hand, the LF model is constructed using a coarser 16-element Timoshenko beam discretization for which the three ballast stiffness parameters are replaced by their equivalent average value and the left and right rail mass parameters are replaced by their average value.
Specifically, we introduce the LF simplification operator
\begin{equation}
    \mathcal{S}(\bm{\theta})
    =[\bar{k}_{B}, \bar{k}_{B}, \bar{k}_{B}, E, \bar{M}, \bar{M}]^{\mathrm{T}},
\end{equation}
where
\begin{equation}
  \bar{k}_{B}
  =\frac{k_{B,l}+k_{B,m}+k_{B,r}}{3},
  \quad
  \bar{M}
  =\frac{M_l+M_r}{2},
\end{equation}
so that the corresponding LF response can be written as \( \mathbf{y}_{\mathrm{LF}} =\mathcal{M}_{16}(\mathcal{S}(\bm{\theta})) \), where \(\mathcal{M}_{16}(\cdot)\) denotes the 16-element model.
Thus, the LF model removes the spatial heterogeneity in the ballast stiffness and the asymmetry in the rail masses.
As a result, the LF model is computationally cheaper and captures the dominant response trend, while the HF model provides a more detailed and accurate representation of the system dynamics.

For the LF dataset \(\mathcal{D}_{\mathrm{LF}}=\{(\bm{\theta}^{(i)},\mathbf{y}_{\mathrm{LF}}^{(i)})\}_{i=1}^{N_{\mathrm{LF}}}\), the uncertain parameter vectors are independently sampled as \(\bm{\theta}^{(i)} \sim \mathcal{U}(0.5,1.5)^6 \) , \(i=1,\ldots,N_{\mathrm{LF}}\), with \(N_{\mathrm{LF}}=4000\).
For each sampled parameter vector, the LF response is computed using the simplified 16-element simulator model, \( \mathbf{y}_{\mathrm{LF}}^{(i)} =\mathcal{M}_{16} \left( \mathcal{S}(\bm{\theta}^{(i)}) \right)\).
Correspondingly, the HF dataset \(\mathcal{D}_{\mathrm{HF}}=\{(\bm{\theta}^{(j)},\mathbf{y}_{\mathrm{HF}}^{(j)})\}_{j=1}^{N_{\mathrm{HF}}}\)  is generated by sampling another set of parameter vectors, \( \bm{\theta}^{(j)} \sim \mathcal{U}(0.5,1.5)^6\) , \(j=1,\ldots,N_{\mathrm{HF}},\) with \(N_{\mathrm{HF}}=400\).
The HF response is obtained from the full 48-element model, \( \mathbf{y}_{\mathrm{HF}}^{(j)} = \mathcal{M}_{48} \left( \bm{\theta}^{(j)} \right)\).
For both datasets, the responses are recorded for three sensors over \(n_t=192\) time steps, resulting in an output dimension of \( W = 3 \times 192 = 576 \).

\begin{figure}[htbp]
    \centering
    \includegraphics[width=\textwidth]{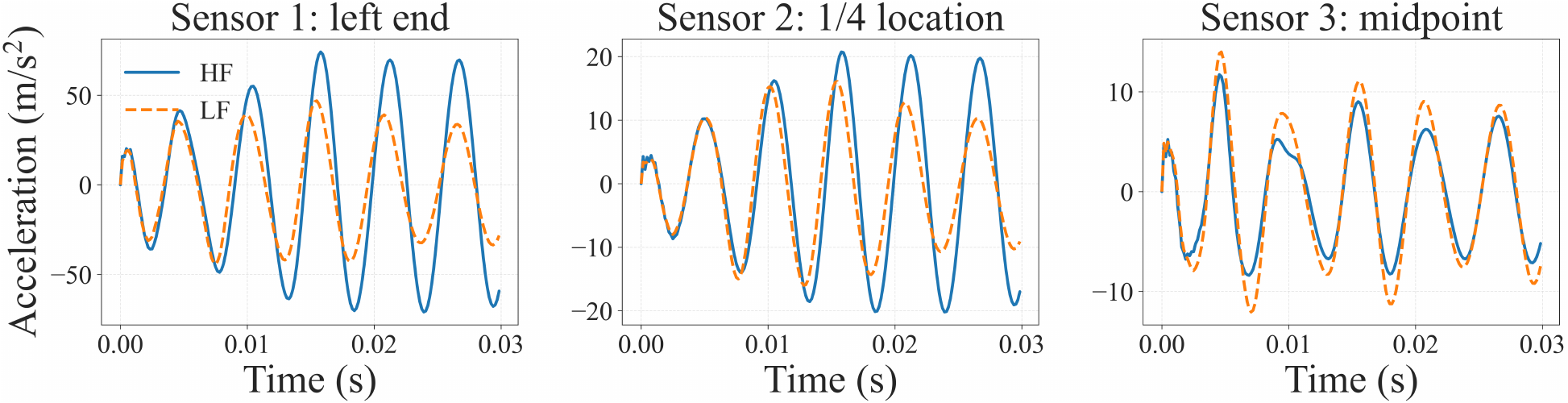}
    \caption{Comparison of LF and HF acceleration responses at three sensor locations for the rail--sleeper--ballast system.}
    \label{fig:lf-hf}
\end{figure}

\Cref{fig:lf-hf} compares the LF and HF acceleration responses at the three sensor locations for the same underlying realization of the physical parameter vector \(\bm{\theta}\).
Clear discrepancies are observed between the LF and HF responses in both amplitude and phase, which arise primarily from the coarser spatial discretization of the LF model and the averaging of the ballast stiffness and rail mass parameters. The comparison therefore shows that the LF model preserves useful information about the underlying dynamic behavior but remains systematically biased, motivating the proposed MF framework to learn a probabilistic correction from LF to HF responses.

\subsubsection{Training of the low-fidelity probabilistic model}\label{sec:rail_pretrain} After constructing the LF and HF datasets, the proposed multi-fidelity learning procedure starts from pretraining a probabilistic LF response model.
The LF model is first trained using the LF-only dataset \(\mathcal{D}_{\mathrm{LF}} \).
Thus, the LF model aims to approximate the conditional distribution \(p(\mathbf{y}_{\mathrm{LF}} \mid \bm{\theta})\), \(\mathbf{y}_{\mathrm{LF}} \in \mathbb{R}^{576}\).
Following the flow architecture introduced in \Cref{sec:flow-architecture}, the LF probabilistic model is constructed using a surjective flow with seven transformation layers, combining both bijective and surjective layers.
The dimensionalities of the successive layers are set as \( [576,\allowbreak 576,\allowbreak 576,\allowbreak 576,\allowbreak 10,\allowbreak 10,\allowbreak 10].\) The first four layers are dimension-preserving affine masked coupling transformations.
The fifth layer is implemented as a masked coupling inference funnel, which reduces the response dimension from \(W=576\) to \(Q=10\). Accordingly, this layer retains a 10-dimensional latent representation and treats the remaining 566 coordinates as the discarded block.
The final two layers are dimension-preserving masked coupling transformations operating in the 10-dimensional latent space. Alternating coordinate permutations are introduced between successive transformations to improve mixing across dimensions, and the final latent variable is assigned a 10-dimensional standard Gaussian base distribution.
The discarded block is modeled using the conditional diagonal Gaussian decoder \labelcref{eq:conditional_decoder}.
The corresponding vectors of conditional means and log-standard deviations are parameterized by a two-headed multi-layer perceptron with two hidden layers of 64 neurons each.
Similarly, the masked coupling transformations \labelcref{eq:mcf-forward} are also parameterized by multilayer perceptrons with two hidden layers of 64 neurons.
    
For training, the model parameters are optimized using the Adam optimizer with a fixed learning rate of \(10^{-4}\). The maximum number of training iterations is set to \(5000\).
A batch size of 128 is used in training. After training, predictions are obtained by Monte Carlo sampling from the learned conditional distribution. Specifically, latent samples are drawn from the Gaussian base distribution and then mapped through the learned inverse conditional flow to generate samples of the LF response. The predictive mean and CIs of the mean are then estimated from these generated response samples.

\begin{figure}[htbp]
    \centering
    \includegraphics[width=\textwidth]{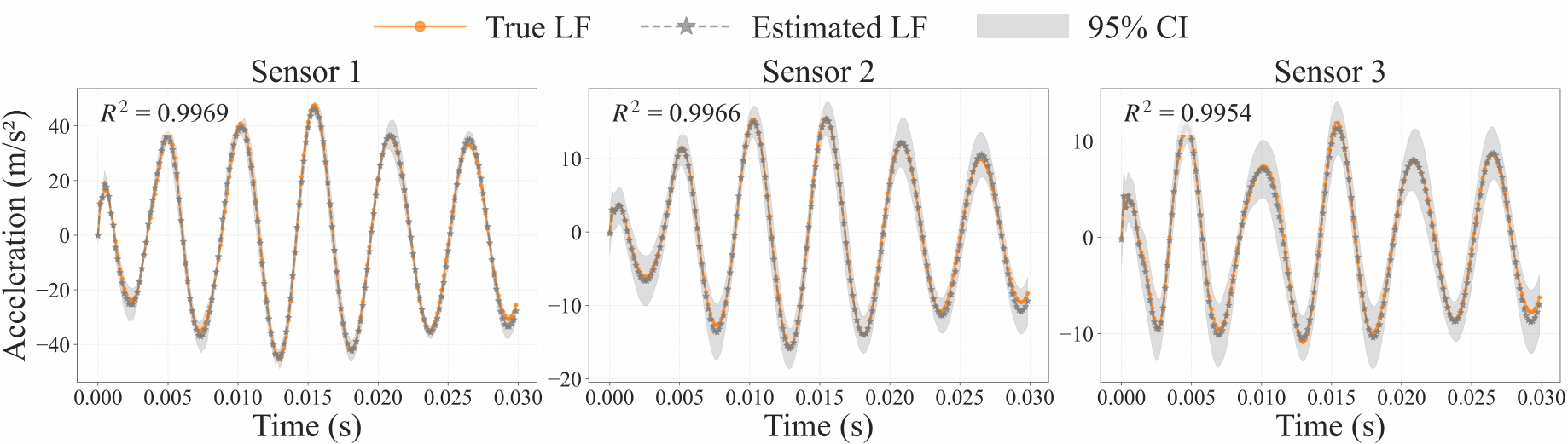}
    \caption{Predictive performance of the trained LF generative model for the rail--sleeper--ballast system.}
    \label{fig:lf-generation}
\end{figure}

\Cref{fig:lf-generation} illustrates the predictive performance of the trained LF generative model for one representative test sample at the three sensor locations.
The orange curves denote the true LF acceleration responses, while the gray dashed curves represent the predictive mean estimated from 2000 Monte Carlo samples generated by the trained LF model. The shaded regions indicate the corresponding 95\% CI.
Overall, the proposed LF generative model accurately reconstructs the LF dynamic responses at all three sensor locations. The predictive mean closely follows the true LF responses in both amplitude and phase, with coefficient of determination \(R^2\) values of 0.9969, 0.9966, and 0.9954 for sensors 1--3, respectively, indicating that the trained LF model successfully learns the conditional distribution of the LF acceleration response given the uncertain physical parameters. It is also observed that the predictive uncertainty is relatively larger near the peak response amplitudes. This behavior is reasonable because the dynamic response is more sensitive to small variations in phase prediction near local extrema and high-amplitude oscillations. Most importantly, the true LF responses are well embedded within the 95\% CI for all three sensors, suggesting that the learned generative model provides reasonable uncertainty quantification.

\subsubsection{Training of the MF model via transfer learning}\label{sec:rail_MF}

After training the LF generative model, we proceed to train the MF model by transferring the learned LF representation to the HF prediction task. Specifically, the trained LF model parameters are used as the initial parameters for the MF model, and the model is subsequently fine-tuned using the available HF data. In this example, 400 HF data are used for fine-tuning. The same flow architecture as the LF model is adopted for the MF model to ensure consistent parameter transfer.
Specifically, the MF model parameters \(\bm{\phi}_{\mathrm{MF}}\) are initialized from the trained LF model parameters, that is we take \(\bm{\phi}_{\mathrm{MF}}^{(0)}=\bm{\phi}_{\mathrm{LF}}\) to initialize the fine-tuning optimization.
We again use the Adam algorithm with a fixed learning rate of \(10^{-4}\). The MF model is fine-tuned for \(1000\) iterations with a batch size of 64.
After training, HF predictions are obtained via Monte Carlo sampling from the learned conditional distribution. For a given input parameter vector \(\bm{\theta}^{*}\), latent samples are drawn from the Gaussian base distribution and mapped through the inverse conditional flow to generate HF response samples,
\(\mathbf{y}_{\mathrm{HF}}^{(m)}
\sim
p_{\bm{\phi}_{\mathrm{MF}}}
\left(
  \mathbf{y}_{\mathrm{HF}}
  \mid
  \bm{\theta}^{*}
\right),
m=1,\ldots,M .
\)

\begin{figure}[htbp]
    \centering
    \includegraphics[width=\textwidth]{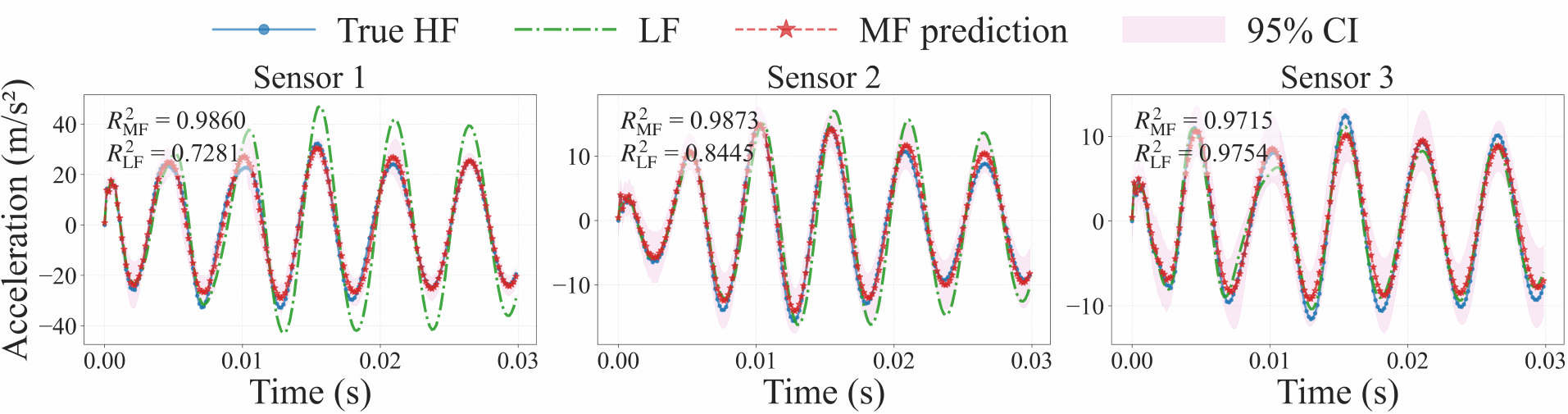}
    \caption{Predictive performance of the fine-tuned MF generative model for the rail--sleeper--ballast system.}
    \label{fig:HF-generation}
\end{figure}

\Cref{fig:HF-generation} shows the predictive performance of the fine-tuned MF generative model for one test sample. The blue curves denote the true HF acceleration responses, the green dashed curves denote the corresponding LF responses, and the red dashed curves denote the MF predictive mean estimated from 2000 generated samples. The shaded regions represent the 95\% CI.
In each subfigure, \(R^2_{\mathrm{MF}}\) denotes the coefficient of determination between the MF predictive mean and the true HF response, while \(R^2_{\mathrm{LF}}\) denotes the coefficient of determination between the LF and true HF responses.

It can be observed that the LF responses provide a useful baseline but exhibit clear discrepancies from the true HF responses, especially for sensors 1 and 2.
After transfer learning and fine-tuning with limited HF data, the MF model substantially improves the prediction accuracy. For sensor 1, the \(R^2\) value increases from \(0.7281\) for the LF response to \(0.9860\) for the MF prediction. For sensor 2, the \(R^2\) value improves from \(0.8445\) to \(0.9873\), demonstrating that the proposed MF model effectively corrects the LF bias and accurately recovers the HF dynamic response.
For sensor 3, the LF response is already close to the HF response, with \(R^2_{\mathrm{LF}}=0.9754\). Therefore, the MF prediction achieves a comparable \(R^2_{\mathrm{MF}}=0.9715\), indicating that limited correction is required at this sensor location. Overall, the MF predictive mean agrees well with the true HF response across all three sensors. The true HF responses are generally well covered by the 95\% CI, suggesting that the fine-tuned MF model provides not only accurate HF predictions but also meaningful uncertainty estimates. 

To further evaluate the effectiveness of the transfer-learning strategy, an additional HF-only model is trained from scratch using the same 400 HF training samples, without initializing from the pretrained LF model. This HF-only model uses the same flow architecture as the proposed MF model, so that the comparison focuses on the effect of LF pretraining rather than model capacity.
The prediction accuracy is quantified using both \(R^2\) and the relative \(\ell_2\) error defined as
\begin{equation}
\label{eq:rel-l2}
\mathrm{Relative\; \ell_{2}\; error}
=\frac{\lVert \mathbf{y}_{\mathrm{pred}} - \mathbf{y}_{\mathrm{true}} \rVert_{2}}
{\lVert \mathbf{y}_{\mathrm{true}} \rVert_{2}},
\end{equation}
where \(\mathbf{y}_{\mathrm{pred}}\) and \(\mathbf{y}_{\mathrm{true}}\) denote the predicted and true HF responses, respectively. A larger \(R^2\) value and a smaller relative \(\ell_2\) error indicate better predictive performance.

\begin{figure}[htpt]
  \centering
  \includegraphics[width=\textwidth]{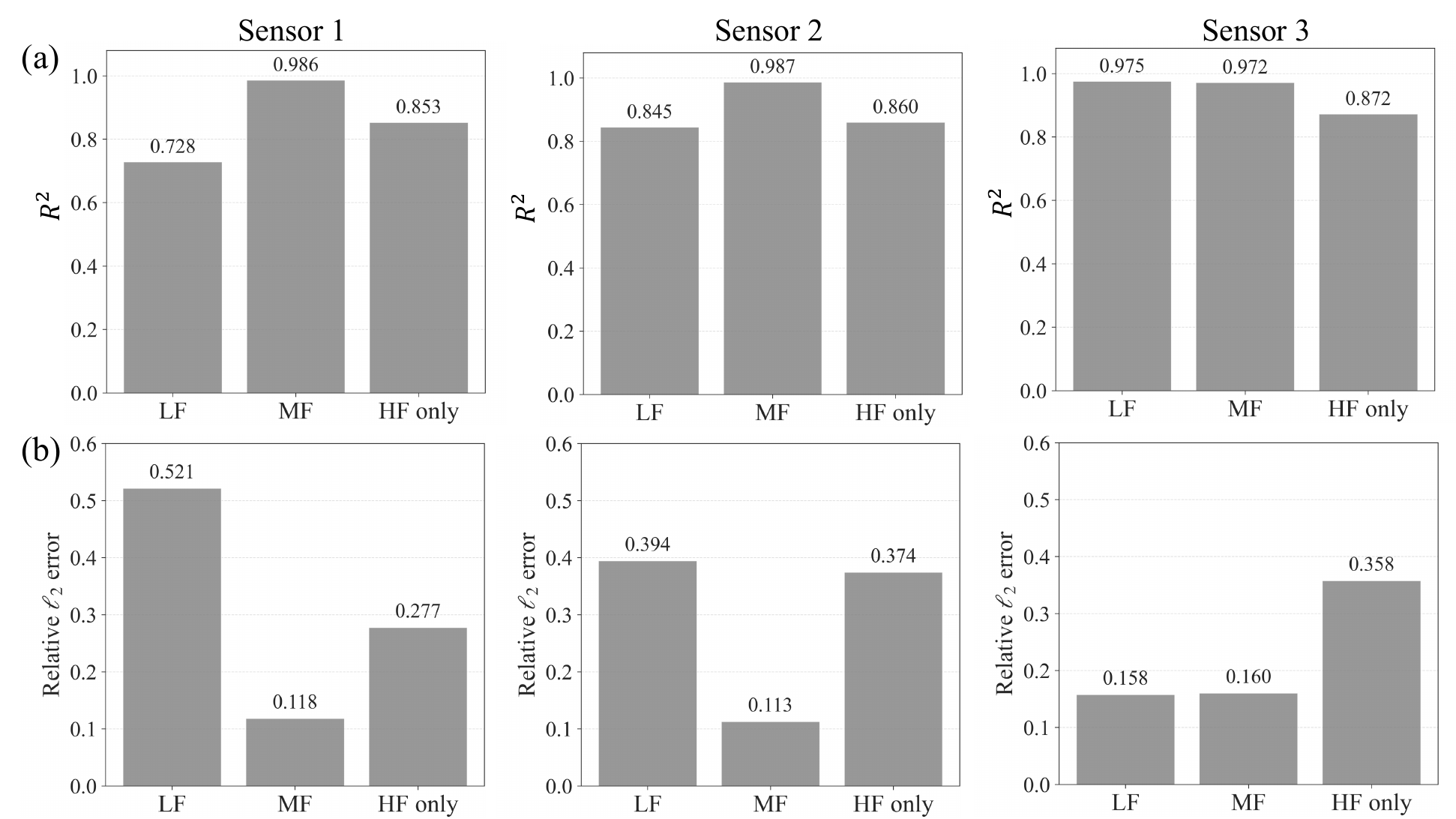}
  \caption{Quantitative comparison of the LF prediction, the MF prediction with LF pre-training, and the HF-only prediction for a representative example of the rail--sleeper--ballast system.}
  \label{fig:rail_metric}
\end{figure}

As shown in \Cref{fig:rail_metric}, the proposed MF model with LF pretraining achieves the best overall performance among the three approaches. For sensor 1, the MF model improves the \(R^2\) value from 0.728 for the LF prediction and 0.853 for the HF-only model to 0.986. Meanwhile, the relative \(\ell_2\) error is reduced from 0.521 for LF and 0.277 for HF-only to 0.118. Although the HF-only model improves upon the LF prediction, its accuracy remains lower than that of the proposed MF model. A similar trend is observed for sensor 2, where the MF model achieves the highest \(R^2\) value of 0.987 and the lowest relative \(\ell_2\) error of 0.113. In contrast, the HF-only model only marginally improves over the LF prediction for sensor 2, suggesting that training a high-dimensional generative model directly from limited HF data is challenging.
For sensor 3, the LF prediction is already close to the HF response, with an \(R^2\) value of 0.975 and a relative \(\ell_2\) error of 0.158, and the MF model provides comparable performance with \(R^2=0.972\) and relative \(\ell_2\) error of 0.160. However, the HF-only model performs noticeably worse, with \(R^2=0.872\) and relative \(\ell_2\) error of 0.358, in line with its performance for sensors 1 and 2.

These comparisons clearly highlight the benefit of transfer learning in the proposed MF framework. Instead of learning the HF response distribution from scratch using only a limited number of HF samples, the MF model starts from the pre-trained LF model, which has already learned the dominant dynamic response patterns from abundant LF data. Fine-tuning then only needs to adapt this learned representation to the HF response space, resulting in improved data efficiency and more accurate prediction when HF simulations are expensive and scarce.

\subsection{Example 2: Reinforced-concrete slab}\label{sec:slab}
\begin{figure}[ht]
    \centering
    \includegraphics[width=\textwidth]{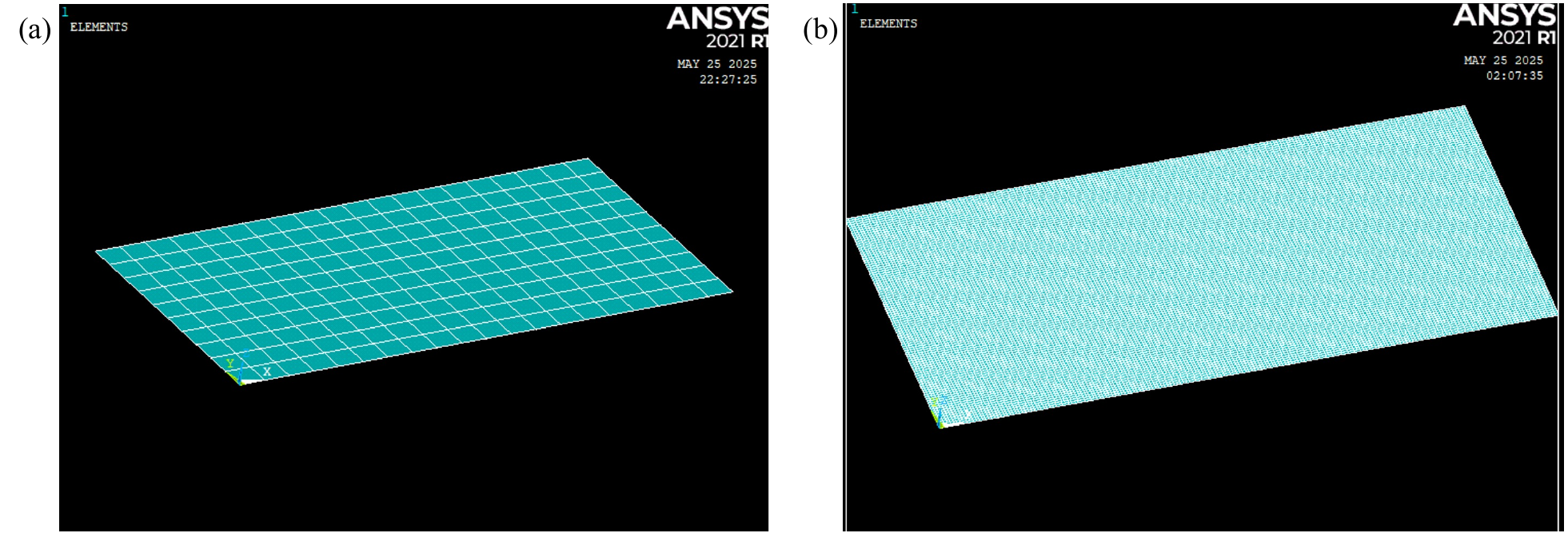} 
    \caption{FE model of a reinforced-concrete slab: (a) LF model; (b) HF model.}
    \label{fig:slab}
\end{figure}

In the second example we consider the modeling of the response of a reinforced concrete slab structure. The quantity of interest is the time-series acceleration response of the slab under ambient excitation.
We consider an elastic structural system with $n$ degrees of freedom, governed by the dynamic equation of motion in \Cref{eq:motion_eq}.
The excitation $\mathbf{f}(t)$ is modeled as a base-acceleration time history consistent with zero-mean, band-limited Gaussian white noise to simulate ambient vibration.
The reinforced concrete slab is modeled using the commercial FE software \textsc{ANSYS} as shown in \Cref{fig:slab}.
The slab is partitioned into nine subsections, and the Young's modulus of each subsection is characterized by a single model parameter, resulting in a total of nine model parameters.
Specifically, each parameter represents the relative deviation in Young’s modulus of the corresponding component with respect to its nominal value, that is,
\begin{equation*}
  \theta_i = \frac{E_i - E_i^{\mathrm{nor}}}{E_i^{\mathrm{nor}}}, \quad i = 1,2,\dots,9,
\end{equation*} where $E_i$ and $E_i^{\mathrm{nor}}$ denote the actual and nominal Young’s moduli, respectively.
All the input parameters are assumed to follow a uniform distribution, $\theta_i \sim \mathcal{U}(-0.3,\,0.3)$.

The two fidelity levels are defined by their mesh discretization, which directly controls the trade-off between cost and accuracy.
Namely, the LF model uses a coarse discretization with a characteristic element size of 0.05~m.
These simulations are inexpensive, allowing for broad coverage of the parameter space.
The HF model uses a fine discretization with a characteristic element size of 0.005~m.
These simulations are computationally costly but provide more accurate acceleration response simulations.
Apart from the mesh density and any differences specified in the test scenarios below, both models share the same geometric description and boundary conditions.

To generate the datasets, we employ \textsc{ANSYS}'s ``Transient Structural Analysis'' capabilities.
For a given input parameter vector $\bm{\theta}$, the system matrices $\mathbf{M}(\bm{\theta})$, $\mathbf{C}(\bm{\theta})$ (e.g., using Rayleigh damping), and $\mathbf{K}(\bm{\theta})$ are assembled. A Gaussian white-noise base-acceleration time history is then created based on the chosen time step $\Delta t$ and simulation duration. 
In the present study, the synthetic Gaussian white-noise excitation is fixed throughout all simulations. Therefore, the simulator is treated as deterministic with respect to the input parameters considered in the FE model, namely the geometric and material properties.
A transient analysis is then performed using the Newmark method, and the time-series acceleration response is recorded at a sensor located at the geometric center of the slab.
This process is repeated for all parameter samples to generate the $\mathcal{D}_{\text{LF}}$ and $\mathcal{D}_{\text{HF}}$ datasets.

To evaluate the robustness of the multi-fidelity surrogate, two distinct scenarios are considered: strong correlation between simulated LF and HF signals (Case 1), and weak correlation between signals (Case 2).
In Case 1, the LF and HF models differ only in their mesh density (0.05~m vs.\ 0.005~m).
All other modeling settings, including the external excitation $\mathbf{f}(t)$, are identical, so that the resulting LF and HF acceleration responses exhibit a strong correlation, and discrepancies are dominated by spatial discretization errors.
Case 2 introduces a more significant discrepancy between the models: In addition to differences in meshing, the LF and HF simulations are subjected to different external excitations.
Let \(\mathbf{f}_{\mathrm{LF / HF}}\) denote the LF/HF excitation; then, we choose \(\mathbf{f}_{\mathrm{HF}}(t) = \mathbf{f}_{\mathrm{LF}}(t) ( 1 + F_{\Delta} \bigr)\) with \(F_{\Delta} = 0.6\), which leads to weakened correlation between the LF and HF responses and a more challenging and realistic transfer learning scenario.

\subsubsection{Case 1: strong correlation between HF and LF data}
\label{sec:case1}

\begin{figure}[ht]
    \centering
    \includegraphics[width=0.85\textwidth]{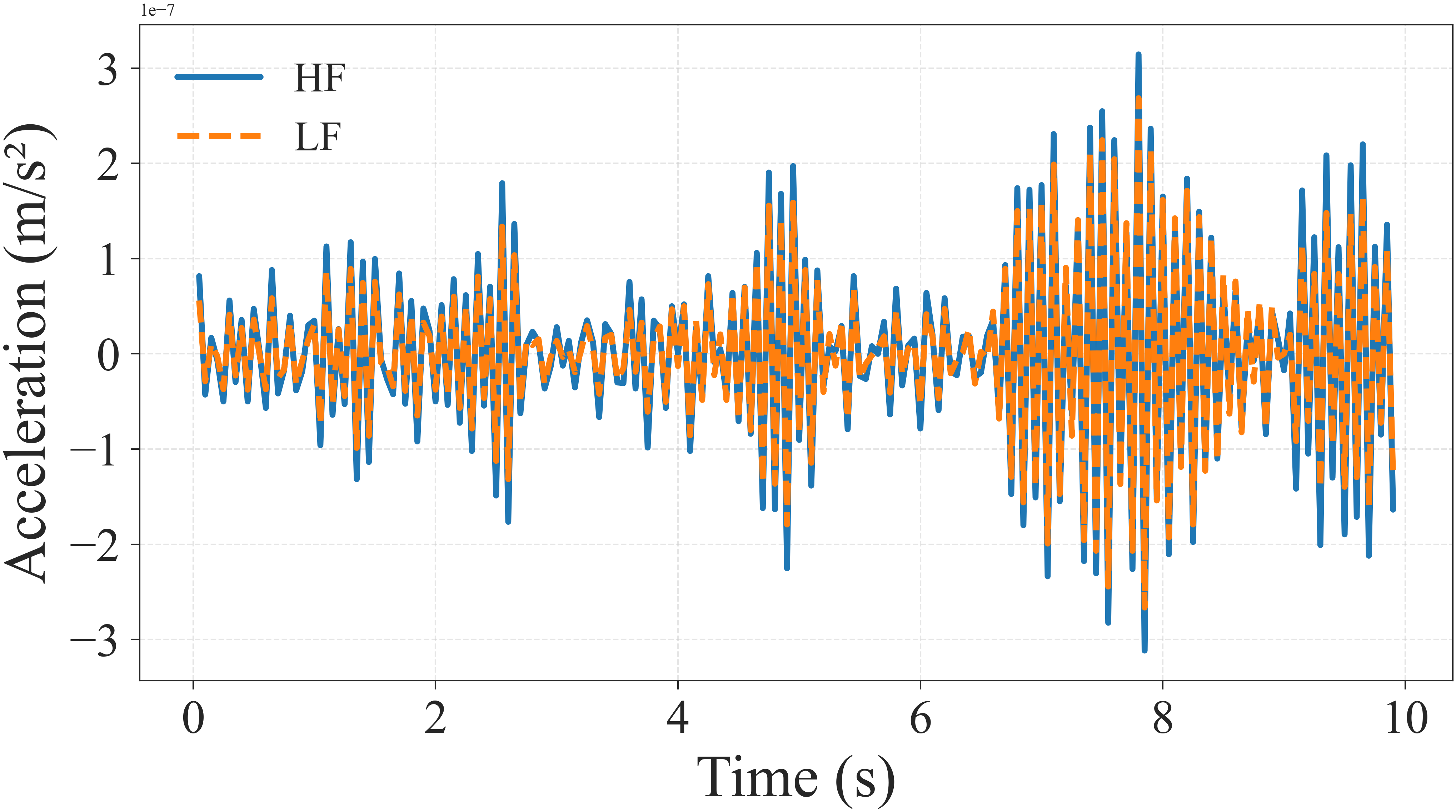} 
    \caption{Example of simulated LF and HF acceleration responses for Case 1.}
    \label{fig:case1_LF_HF}
\end{figure}

To emulate a realistic data-imbalance setting, we generate 1000 LF and 200 HF acceleration time-series data pairs. All \textsc{ANSYS} simulations were performed on a computer equipped with an Intel Core i7-10510U processor operating at 1.8~GHz and 8~GB of memory. The total data-generation time was approximately 4.52~hours for the LF simulations and 13.47~hours for the HF simulations.
The responses are sampled at 20 Hz over a 10 seconds window. \Cref{fig:case1_LF_HF} presents an example with strong LF–HF correlation: the LF signal shows modest amplitude/phase deviations from the HF reference, and both traces follow similar temporal patterns.

As discussed in \cref{sec:method}, the LF model is first trained using 980 LF data pairs, with the remaining samples used for validation. The surjective flow model consists of seven layers, with dimensionalities \([200,\allowbreak 200,\allowbreak 200,\allowbreak 200,\allowbreak 10,\allowbreak 10,\allowbreak 10]\). The first four layers are dimension-preserving, while the fifth layer performs a dimensionality reduction from \(W=200\) to \(Q=10\), corresponding to a reduction factor of 95\%. The final two layers operate bijectively. The rationale behind this architecture is that the dataset is fully processed through bijective mappings before data compression, ensuring that sufficient structure is extracted prior to dimensionality reduction. Subsequent bijective refinement after reduction is aimed to balance expressiveness with efficiency.

For training, optimization is performed using the Adam algorithm \cite{jais2019adam} with a fixed learning rate of \(10^{-4}\).
A batch size of 64 is used throughout, and training is performed for 1000 iterations.
After training, predictions are obtained by Monte Carlo sampling from the learned conditional distribution. For each input, latent samples are first drawn from the Gaussian base distribution and then mapped through the learned inverse conditional flow to generate samples of the output. The predictive mean and CIs are subsequently estimated from these generated samples.

\begin{figure}[ht]
    \centering
    \includegraphics[width=\textwidth]{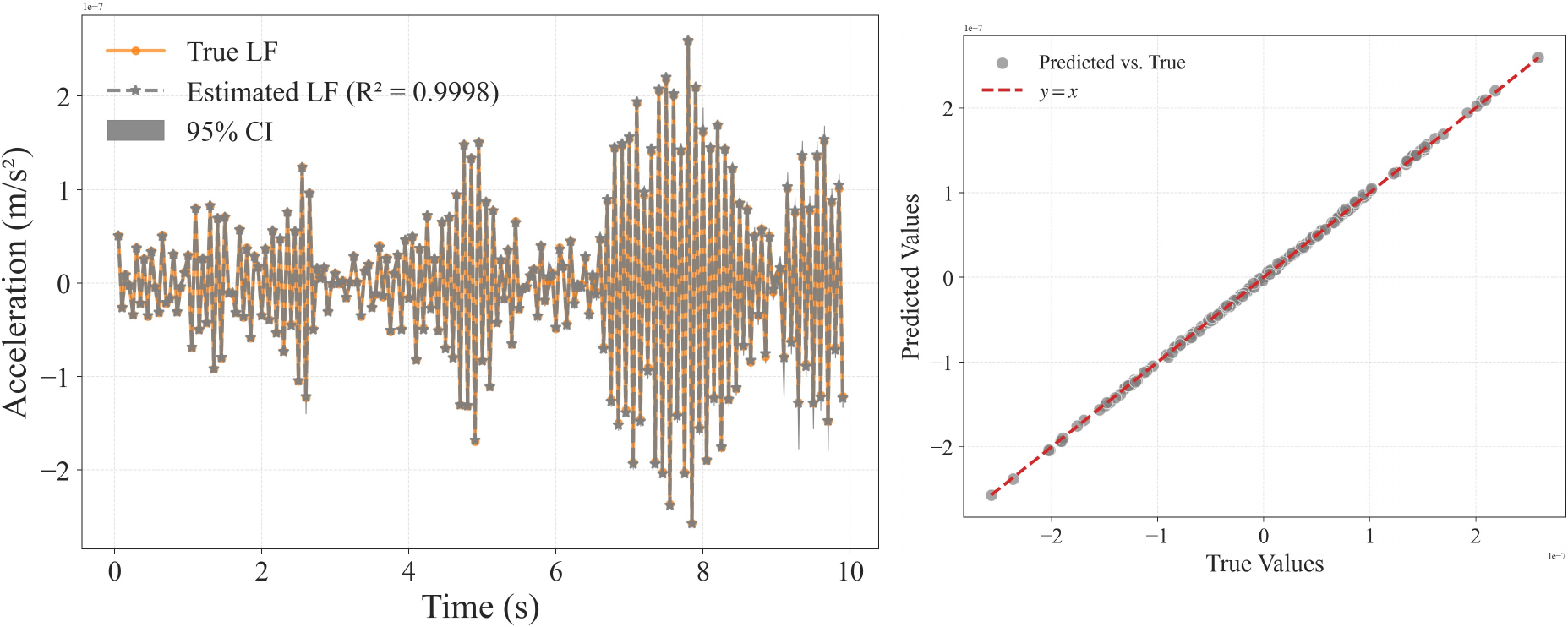} 
    \caption{Validation of the LF model for Case 1.}
    \label{fig:case1_LF_vali}
\end{figure}

\Cref{fig:case1_LF_vali} presents the validation performance of the trained LF model. The left panel compares the reconstructed LF response against the true LF time history for a representative case. The predicted mean response aligns well with the simulated data, and the shaded region indicates the 95\% CI. The narrow width of the CI demonstrates that the model not only captures the correct temporal pattern but also maintains low predictive uncertainty. The coefficient of determination, $R^{2}=0.9998$, further confirms the good agreement between the estimated and true LF responses. The right panel shows a scatter plot of predicted versus true LF values. The points cluster tightly around the reference line $y=x$, indicating highly accurate predictions.

\begin{figure}[ht]
    \centering
    \includegraphics[width=\textwidth]{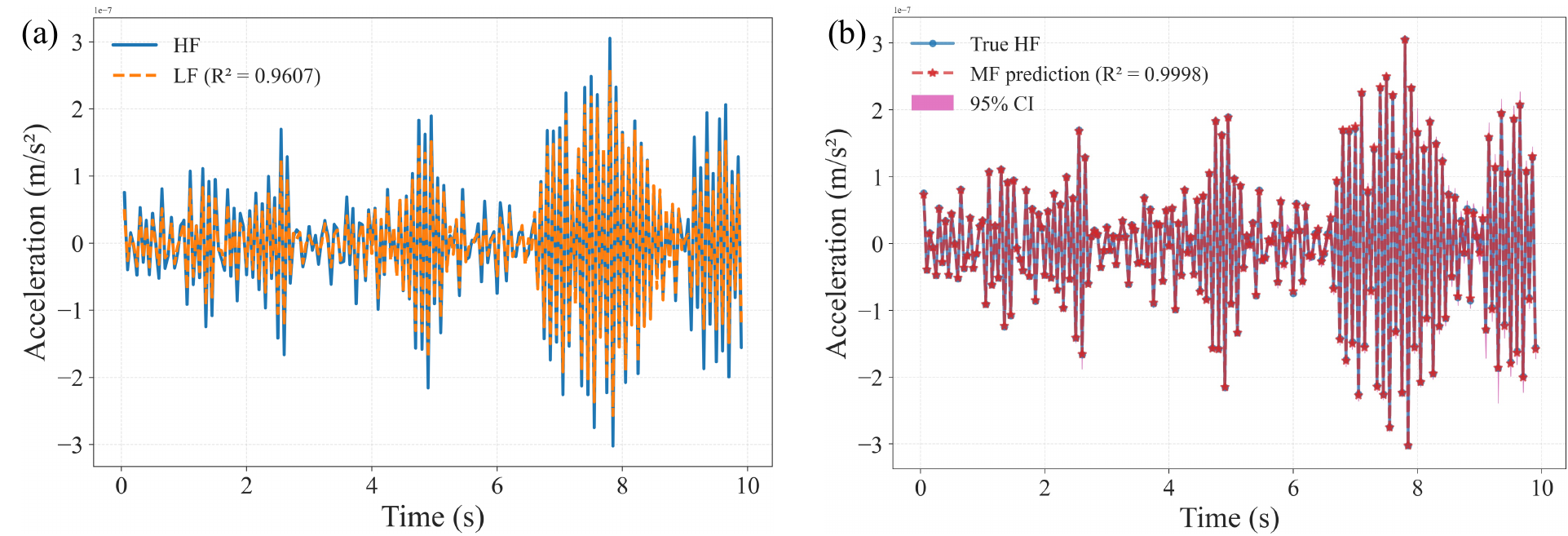} 
    \caption{MF prediction performance on test data \#1 for Case 1: (a) comparison between HF and LF simulated responses; (b) MF predicted mean and 95\% compared against simulated HF response.}
    \label{fig:case1_MF_test1}
\end{figure}

\begin{figure}[ht]
    \centering
    \includegraphics[width=\textwidth]{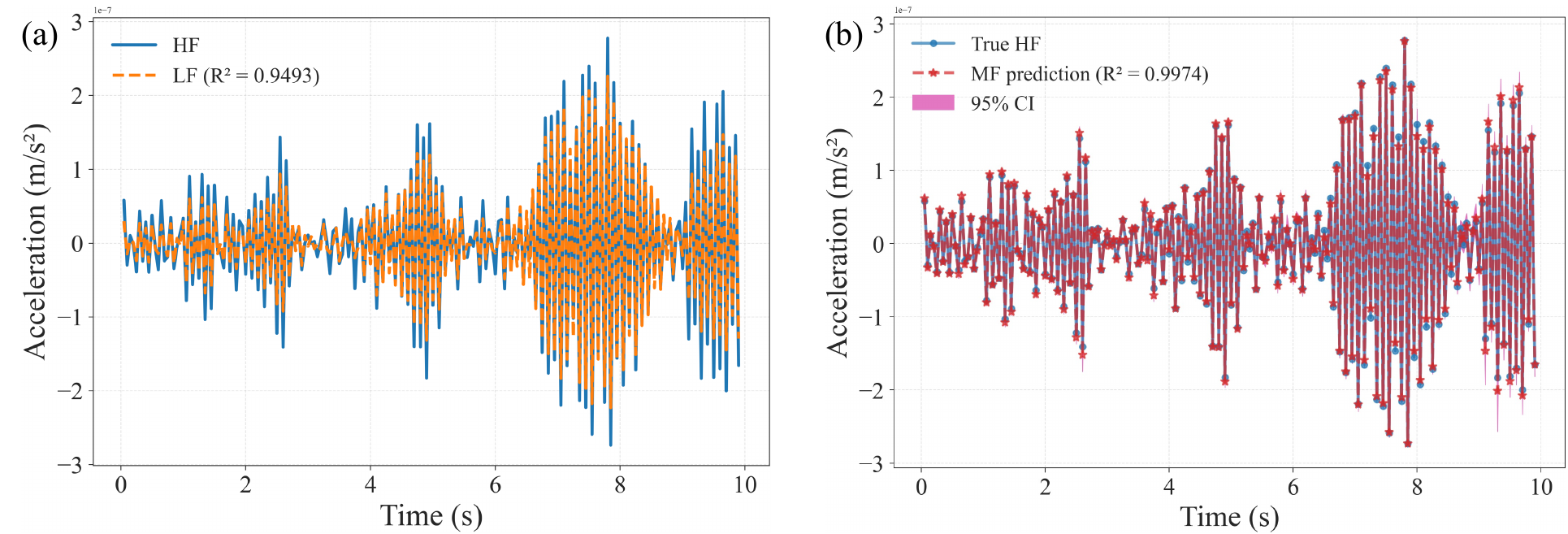} 
    \caption{MF prediction performance on test data \#2 for Case 1: (a) comparison between HF and LF simulated responses; (b) MF predicted mean and 95\% compared against simulated HF response.}
    \label{fig:case1_MF_test2}
  \end{figure}

Next, we proceed to train a MF model building upon the previously trained LF model.
The LF model is used as a pre-trained initialization and subsequently fine-tuned using 180 available HF data pairs, while the remaining 20 HF data pairs are reserved for testing. The MF fine-tuning is performed for 500 iterations with a batch size of 16.
\Cref{fig:case1_MF_test1} and \Cref{fig:case1_MF_test2} show the true HF response and the MF model prediction for two test data pairs.
In both figures, panel (a) compares the simulated HF and LF response for the corresponding inputs, while panel (b) compares the simulated HF response against the MF predicted mean with the associated 95\% CI. For test dataset \#1 (Fig.~\ref{fig:case1_MF_test1}), the LF response reproduces the overall temporal trend of the HF data but deviates in amplitude and phase, particularly in the energetic burst around 7--9~s, leading to a moderate fit of $R^{2}=0.9607$. In contrast, the MF prediction achieves better agreement with the HF response ($R^{2}=0.9998$). The 95\% CI is narrow across most of the time domain, indicating the reliability of predicted responses. For test dataset \#2 (Fig.~\ref{fig:case1_MF_test2}), the LF model again captures the general pattern but exhibits larger deviations at peak responses, reflected in a lower $R^{2}=0.9493$. The MF prediction, however, substantially improves the accuracy ($R^{2}=0.9974$), closely following both the amplitude and phase of the HF data. The uncertainty intervals also remain tight overall, reflecting appropriate predictive capacity over the prediction window.

Overall, the comparison between LF and MF demonstrates a significant improvement achieved by the proposed approach. While LF provides only an approximate proxy to HF dynamics, the MF model, through pre-training on abundant LF data and fine-tuning with limited HF data, learns an underlying representation that preserves critical features of the HF response. This enables accurate reproduction of both global trends and local transients, while providing reliable uncertainty quantification.

\subsubsection{Case 2: weak correlation between HF and LF}
\label{sec:case2}

In the second case, the LF and HF datasets exhibit a weaker correlation compared to Case 1 in \Cref{sec:case1}. This weak correlation is intentionally introduced by increasing the discrepancy between the external excitations applied during LF and HF data generation. Apart from this modification, the modeling procedures remain identical to those in Case 1.

\begin{figure}[H]
    \centering
    \includegraphics[width=0.85\textwidth]{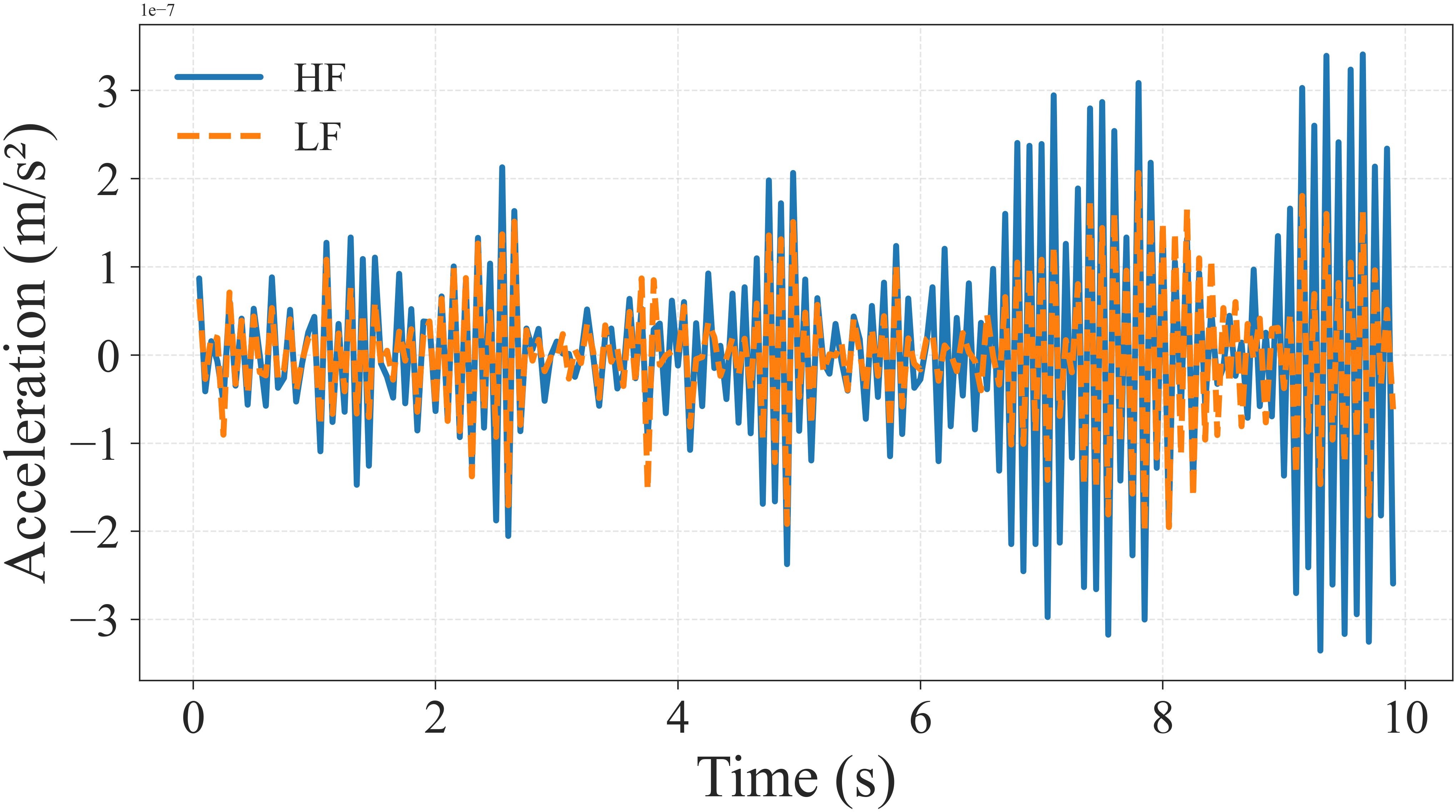} 
    \caption{Example of simulated LF and HF acceleration responses for Case 2.}
    \label{fig:case2_LF_HF}
\end{figure}

\Cref{fig:case2_LF_HF} illustrates a representative comparison between LF and HF accelerations.
Unlike Case 1 (\Cref{fig:case1_LF_HF}), where the LF and HF responses followed a similar pattern with relatively minor amplitude and phase mismatches, here the discrepancy between LF and HF data becomes much more pronounced.
The LF data no longer provides a reliable approximation of the HF dynamics, particularly in high-energy segments where deviations in amplitude and phase accumulate over time.
This intentionally degraded correlation setting poses a more challenging scenario for MF training and transfer learning.
Since the LF data carry less direct information about the HF response, the MF model must rely more heavily on the fine-tuning process with limited HF data to achieve accurate predictions.
As a result, Case 2 serves as a more stringent test of the proposed MF framework, demonstrating its robustness in situations where LF models are of limited fidelity and the LF--HF relationship is highly nonlinear and weakly coupled.

\begin{figure}[H]
    \centering
    \includegraphics[width=\textwidth]{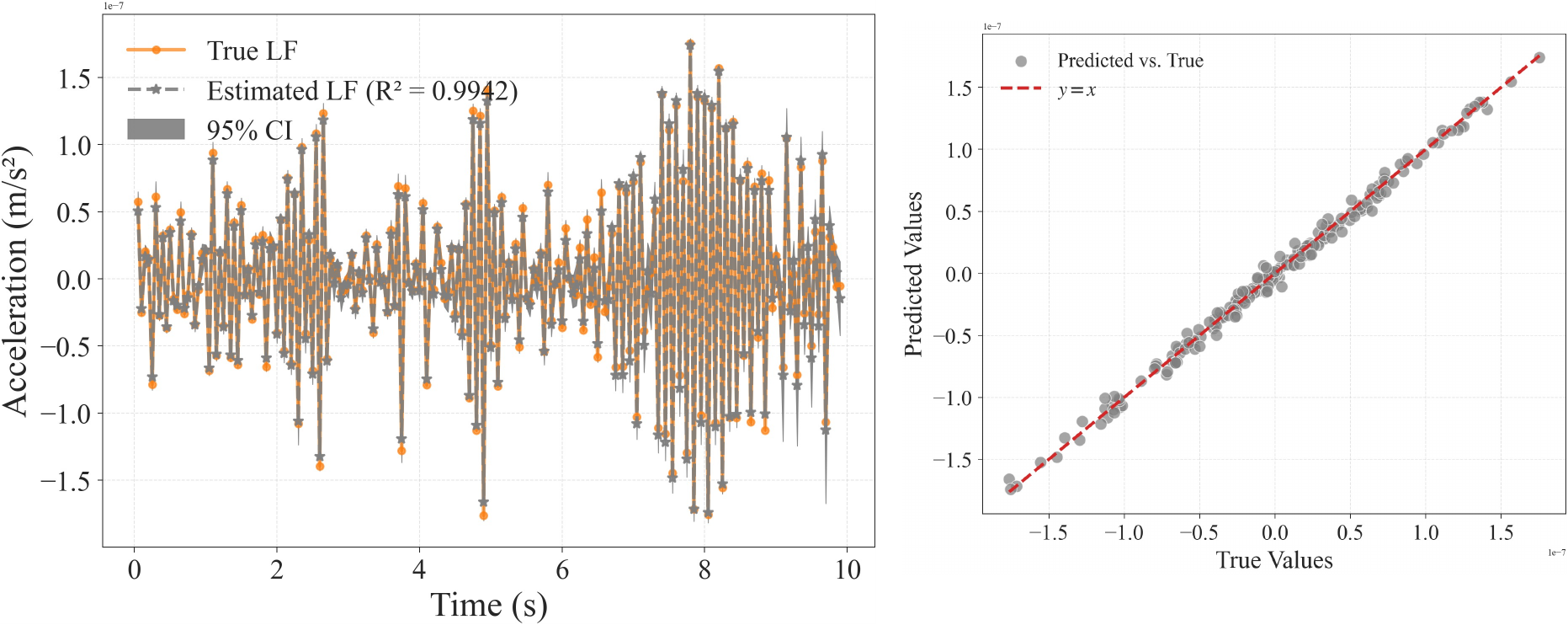} 
    \caption{Validation of the LF model for Case 2.}
    \label{fig:case2_LF_valid}
\end{figure}

We again train the LF model using 980 LF datasets and 20 additional datasets for validation. Each dataset consists of 200 data points sampled over a 10-second duration. To ensure consistency with Case 1, the model architecture is taken to be the same. Specifically, the surjective NF model comprises seven flow layers with dimensionalities [200, 200, 200, 200, 10, 10, 10].
\Cref{fig:case2_LF_valid} presents the validation performance of the trained LF model. On the left, the estimated LF response is compared against the true LF response. The two curves align closely across the full time window, yielding a high $R^{2} = 0.9942$. The uncertainty band remains narrow, indicating that the LF model is also confident in its predictions. On the right, a point-to-point comparison between predicted and true LF data is shown, further confirming the excellent agreement between model predictions and simulated data. The results demonstrate that the LF surjective NF model provides reliable estimates of LF responses.

\begin{figure}[ht]
    \centering
    \includegraphics[width=\textwidth]{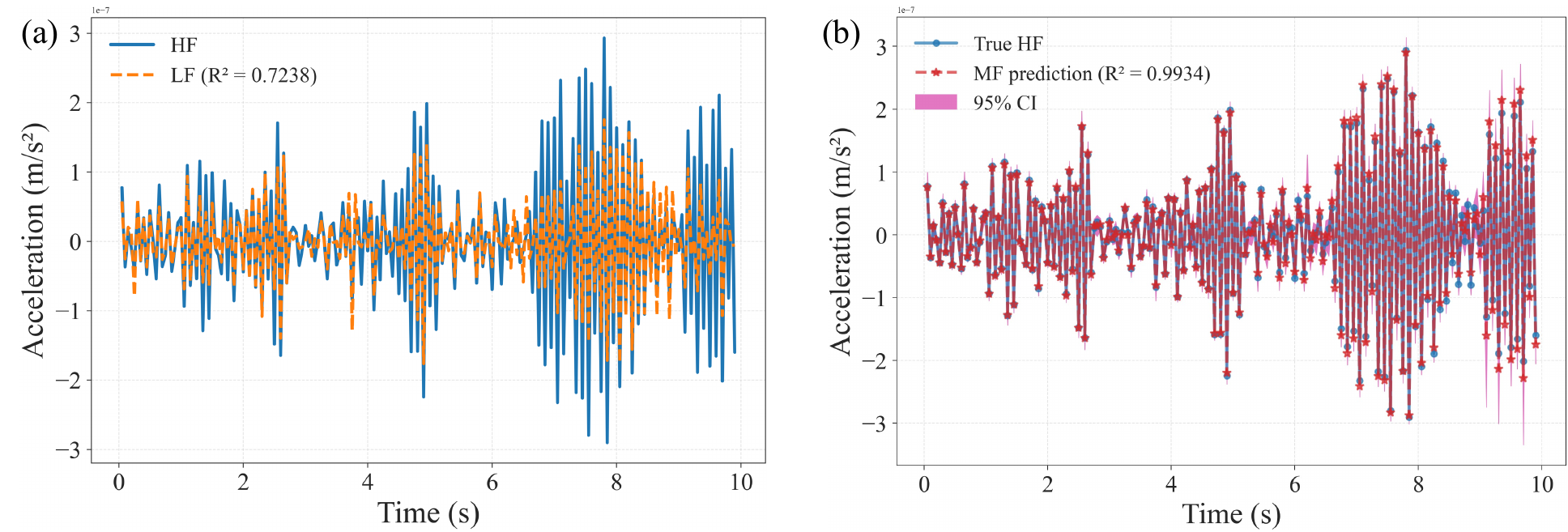} 
    \caption{MF prediction performance on test data \#1 for Case 2: (a) comparison between HF and LF simulated responses; (b) MF predicted mean and 95\% compared against simulated HF response.}
    \label{fig:case2_MF_test1}
\end{figure}

\begin{figure}[H]
    \centering
    \includegraphics[width=\textwidth]{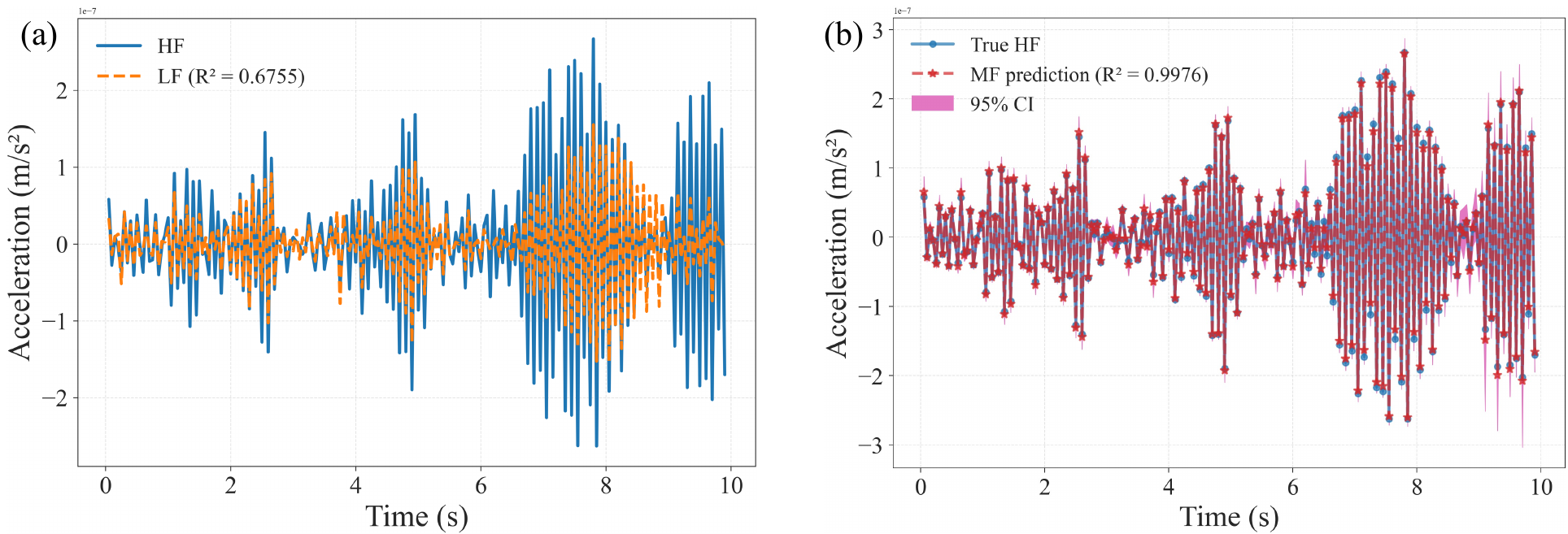} 
    \caption{MF prediction performance on test data \#2 for Case 2: (a) comparison between HF and LF simulated responses; (b) MF predicted mean and 95\% compared against simulated HF response.}
    \label{fig:case2_MF_test2}
\end{figure}

We proceed to fine-tune the pre-trained LF model into the MF model by using 180 HF data pairs, with the remaining 20 HF data pairs are used for testing.
\Cref{fig:case2_MF_test1,fig:case2_MF_test2} illustrate the MF prediction results for two test data pairs. In both figures, panel~(a) compares the simulated HF and LF responses for the corresponding inputs. It can be seen that the LF responses deviate substantially from the HF responses, particularly in amplitude and phase during high-energy segments. This large discrepancy highlights the weak correlation between LF and HF in Case 2 and underscores the challenge of reducing such differences through MF learning.
Panels~(b) in \Cref{fig:case2_MF_test1,fig:case2_MF_test2} present the MF mean predictions and 95\% CIs against the simulated HF responses. Despite the weak LF--HF correlation, the MF predictions align closely with the HF data across the entire time window, with $R^{2}$ values exceeding 0.99 in both cases. The 95\% CIs remain generally narrow, indicating a good prediction reliability. Importantly, the uncertainty widens in the final two seconds of the signal, where the HF responses exhibit stronger fluctuations, indicating that the predictive uncertainty increases in regions where the HF dynamics become more oscillatory and complex. Overall, these results confirm that the MF surjective NF model can successfully overcome weak LF--HF correlations, accurately recovering the HF dynamics while providing uncertainty estimates that adapt to the underlying structural response complexity.  


\subsubsection{Effect of the amount of HF data}

To further evaluate the robustness of the proposed approach, we investigate the effect of the number of HF data pairs used for fine-tuning on the performance of the MF model. Two additional scenarios are considered: (i) training a single NF model using only 180 HF data pairs without LF pre-training, and (ii) training a MF model using only 100 HF data pairs with LF pre-training as discussed in \cref{sec:case1,sec:case2}. 

\begin{figure}[H]
    \centering
    \includegraphics[width=\textwidth]{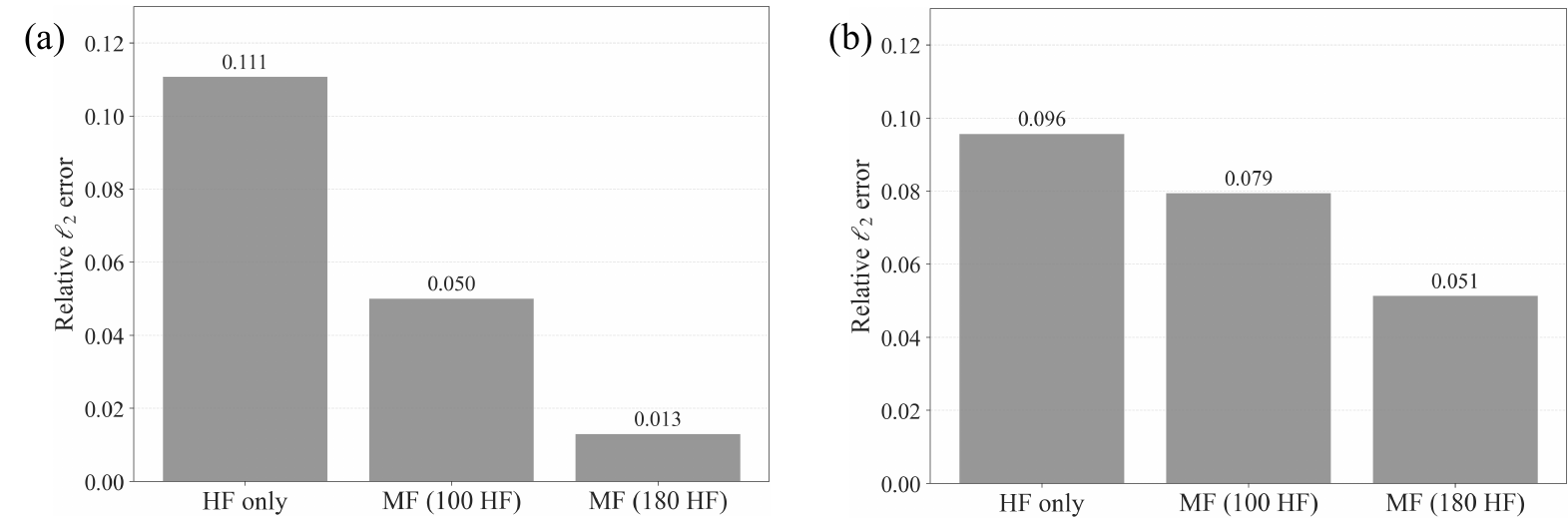} 
    \caption{Effect of different amounts of HF data in Case 1 (strong correlation): (a) Test data \#1; (b) Test data \#2.}
    \label{fig:case1_differ_HF}
\end{figure}

\begin{figure}[H]
    \centering
    \includegraphics[width=\textwidth]{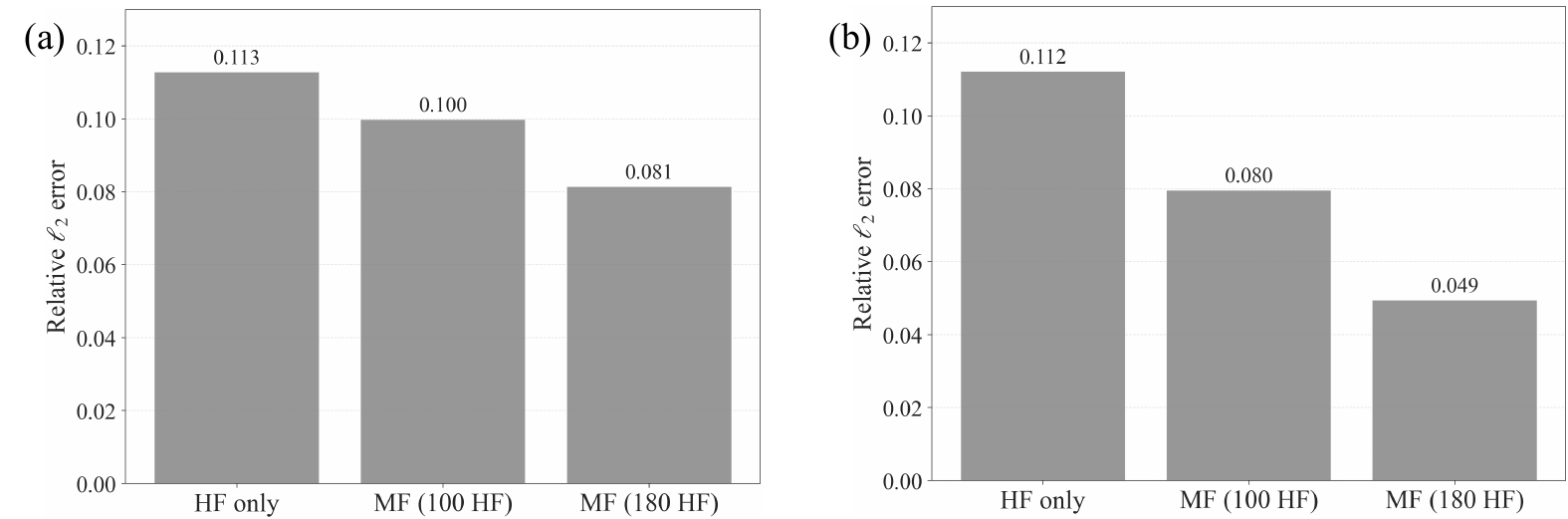} 
    \caption{Effect of different amounts of HF data in Case 2 (weak correlation): (a) Test data \#1; (b) Test data \#2.}
    \label{fig:case2_differ_HF}
  \end{figure}

\Cref{fig:case1_differ_HF,fig:case2_differ_HF} present the results for Case 1 (strong LF–HF correlation) and Case 2 (weak LF–HF correlation), respectively, across three scenarios: HF-only (180 data pairs), MF with 100 HF data pairs, and MF with 180 HF data pairs. In both cases and for both test datasets, the consistent observation is that fine-tuning the LF model with 180 HF datasets yields the lowest relative $\ell_{2}$ errors. By contrast, the HF-only model trained on 180 HF data pairs without LF pre-training consistently performs worst, confirming that the absence of LF information leads to poor overall performance. The MF model trained with 100 HF datasets provides intermediate performance, but still outperforms the HF-only baseline, underscoring the benefit of leveraging LF data through transfer learning.
  
These results illustrate that the integration of LF knowledge is critical for efficient utilization of limited HF data. The MF model not only reduces the required number of HF datasets for training, but also delivers superior accuracy when fine-tuned with sufficient LF information. Furthermore, when the pre-trained LF model is used, incorporating more HF data significantly improves the predictive performance of the MF model. This trend highlights the complementary roles of LF knowledge transfer and sufficient HF information: the LF model provides a strong prior, while additional HF data refines the model to better capture the dynamics of the high-fidelity system.

\subsubsection{Comparison against a conventional multi-fidelity ML approach}
\label{sec:comparison}

To further demonstrate the efficacy of the proposed MF surrogate modeling framework, we conduct a comparison against a traditional non-generative machine learning approach.
In this study, we adopt the TCN as the backbone of the MF framework for comparison \cite{zhong2025multi, zeng2023machine}. TCN extends conventional convolutional neural networks (CNN) to sequential modeling by introducing causal convolutions and dilated filters, which allow the network to capture long-range temporal dependencies while preserving the ordering of time-series data. Owing to their efficiency and accuracy, TCN has been widely applied in time-series forecasting and sequence-to-sequence prediction tasks \cite{hewage2020temporal, fan2023parallel}.

In the MF surrogate setting, the TCN is employed in a manner analogous to the proposed flow-based model. Specifically, a LF surrogate is first trained using abundant LF data, and the pre-trained TCN is then fine-tuned to construct a MF surrogate using limited HF data. The input to the TCN consists of the structural model parameters concatenated with the time step, such that the network predicts point-wise responses across the 200 time steps of interest. Unlike the generative flow-based framework, the TCN produces deterministic predictions without explicit uncertainty quantification. The TCN architecture adopted in this study takes 10 input features (9 model parameters and one time step) and predicts a single output at each time step. The temporal convolutional layers consist of 10 blocks, each with 60 channels, enabling the network to capture complex temporal patterns. A kernel size of 20 allows the receptive field to span longer sequences, while a dropout rate of 0.2 is introduced to mitigate overfitting.
The comparison is carried out for Case 2, in which the LF–HF correlation is weak. This scenario is intentionally chosen because it represents the more challenging setting relative to Case 1 and therefore provides a stricter test of the proposed MF framework. For a fair comparison, the same 1000 LF and 200 HF data are used for training TCN.

\begin{figure}[ht]
  \centering
  \begin{subfigure}[b]{\textwidth} 
    \centering
    \includegraphics[width=\textwidth]{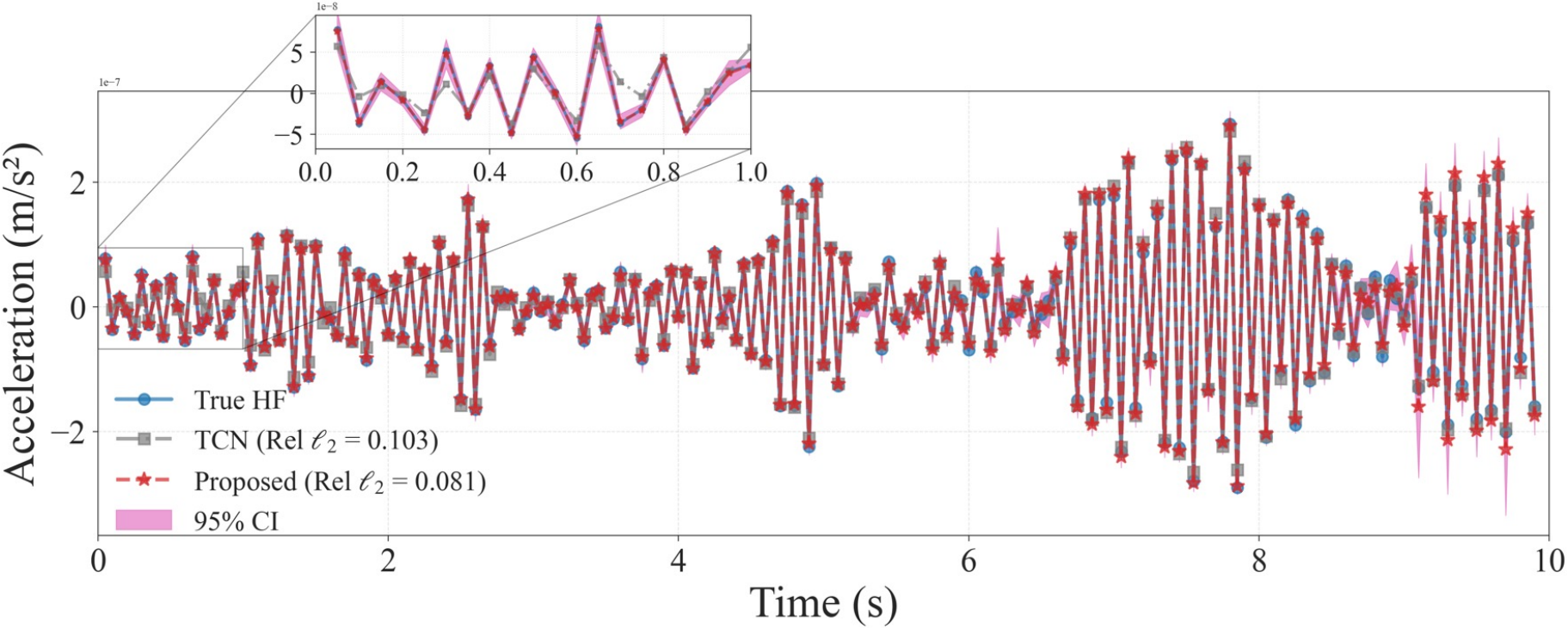}
    \caption{Test \#1}
    \label{fig:comparison_a}
  \end{subfigure}
  \hfill 
  \begin{subfigure}[b]{\textwidth} 
    \centering
    \includegraphics[width=\textwidth]{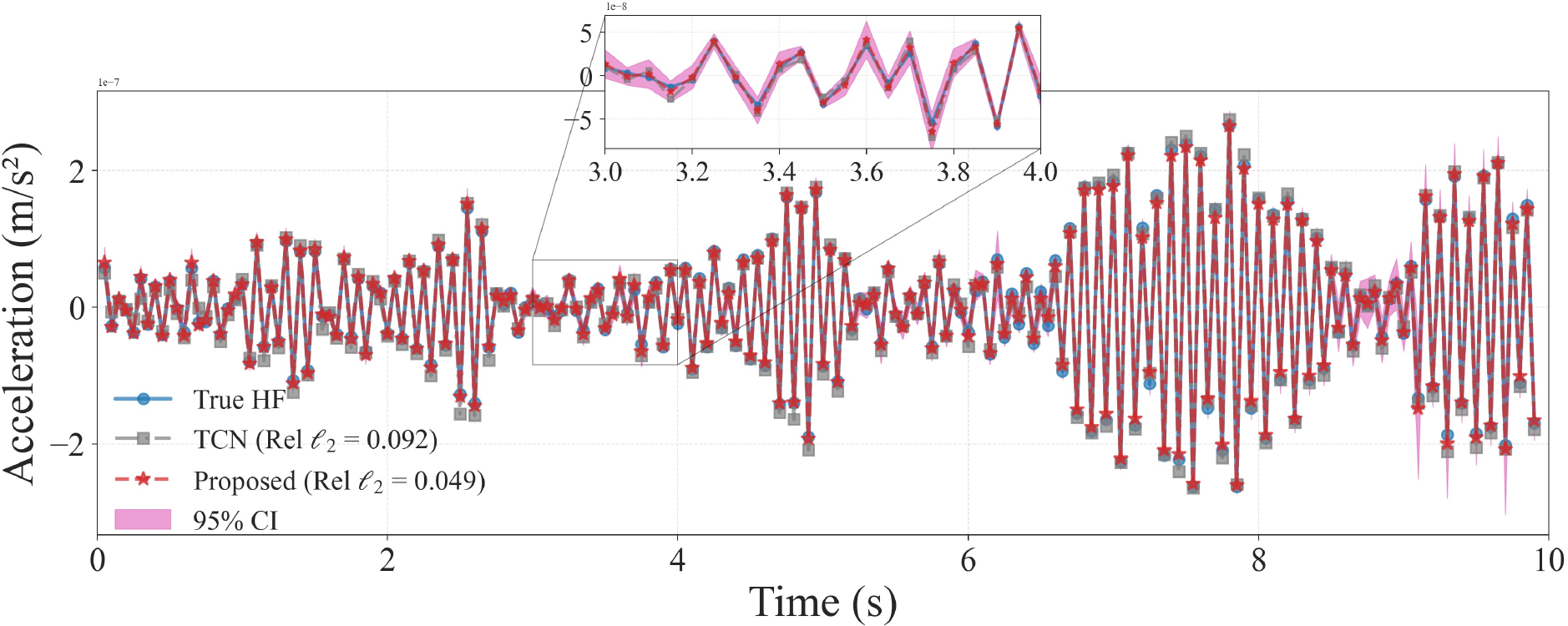} 
    \caption{Test \#2}
    \label{fig:comparison_b}
  \end{subfigure}
  \caption{Comparison between the proposed method (predictive mean and 95\% CI) and the TCN-based MF method for two test data examples.}
  \label{fig:comparison_TCN}
\end{figure}

\Cref{fig:comparison_TCN} compares the prediction performance of the proposed framework against the TCN-based method for two HF test data pairs in Case 2. In both subfigures, the blue circles denote the true HF responses, the grey squares represent TCN predictions, and the red stars correspond to the proposed method. Overall, both methods are able to capture the general dynamic trends of the HF responses. However, the proposed framework demonstrates noticeably improved accuracy, as reflected in the smaller relative $\ell_2$ errors (0.081 and 0.049 for the proposed method, compared to 0.103 and 0.092 for TCN in \Cref{fig:comparison_a} and \Cref{fig:comparison_b}, respectively). These improvements indicate that the proposed approach not only provides more accurate pointwise predictions but also better preserves the fine-scale temporal dynamics. The zoomed-in insets highlight regions where discrepancies are more evident. 

While the TCN-based method exhibits slight phase mismatches and amplitude deviations from the ground truth, the proposed method remains closely aligned with the HF data. This demonstrates the advantage of the proposed probabilistic generative approach in learning complex temporal structures, particularly under weak LF--HF correlation settings where conventional deterministic ML methods such as TCN tend to degrade. Another key advantage of the proposed generative AI–enhanced MF modeling framework is that it produces full predictive probability distributions rather than point estimates as in the case of TCN, as shown in the shaded areas in zoomed-in insets representing 95\% CIs. This capability allows the model to naturally quantify predictive uncertainty, which is critical for structural reliability assessment and risk-informed decision making.
Furthermore, to rigorously quantify predictive accuracy, we computed the average relative $\ell_2$ errors between the MF prediction and the HF true data across 20 distinct, unseen test datasets for both the TCN and the proposed method. The TCN yielded a mean error of 0.091. In contrast, the proposed method achieved a lower average error of 0.0791 (12.7\% reduction), indicates better accuracy and improved generalizability to unseen inputs.

\begin{figure}[H]
    \centering
    \begin{subfigure}[b]{0.48\textwidth} 
        \centering
        \includegraphics[width=\textwidth]{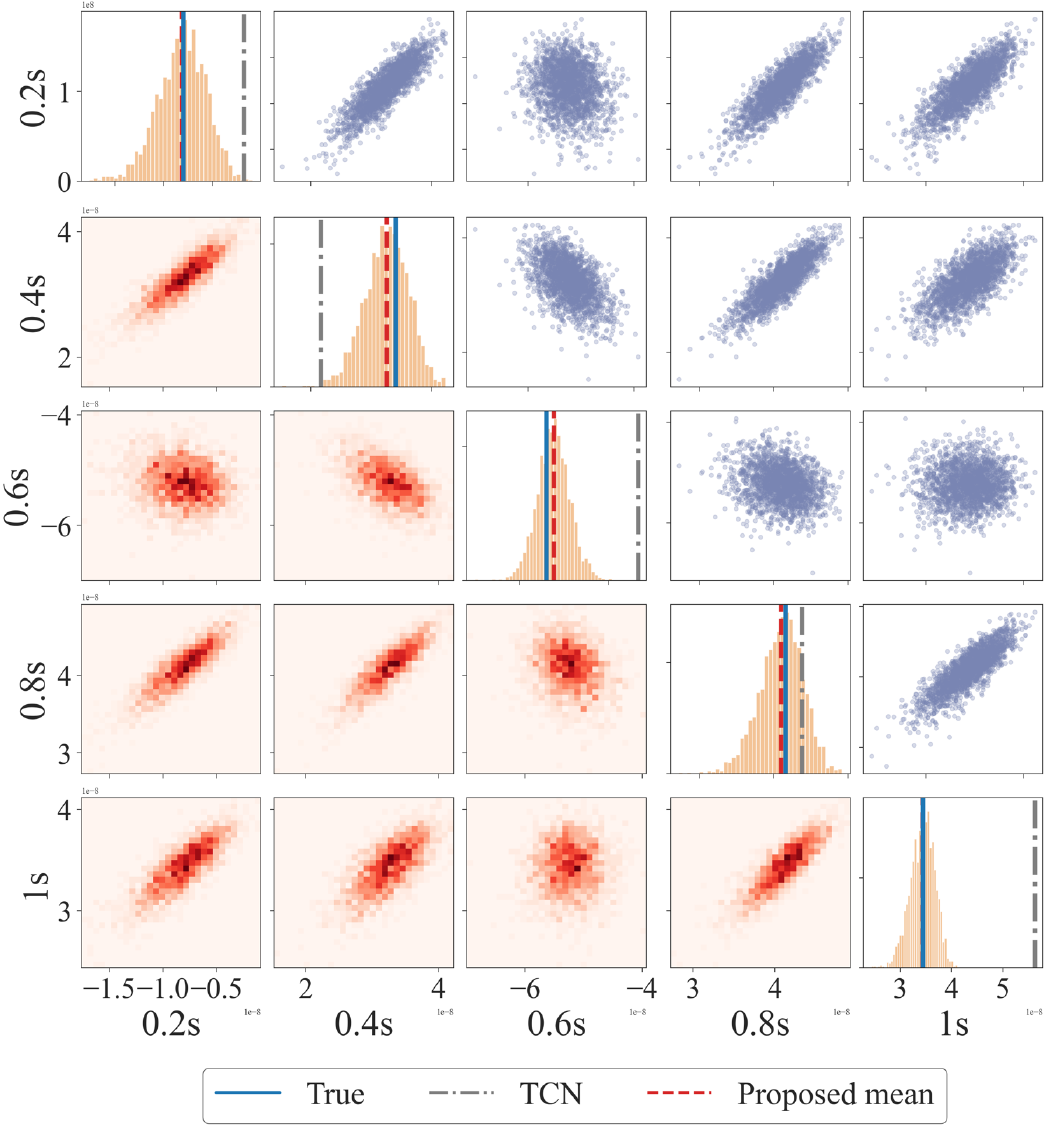}
        \caption{Test \#1}
        \label{fig:dis_a}
    \end{subfigure}
    \hfill 
    \begin{subfigure}[b]{0.48\textwidth} 
        \centering
        \includegraphics[width=\textwidth]{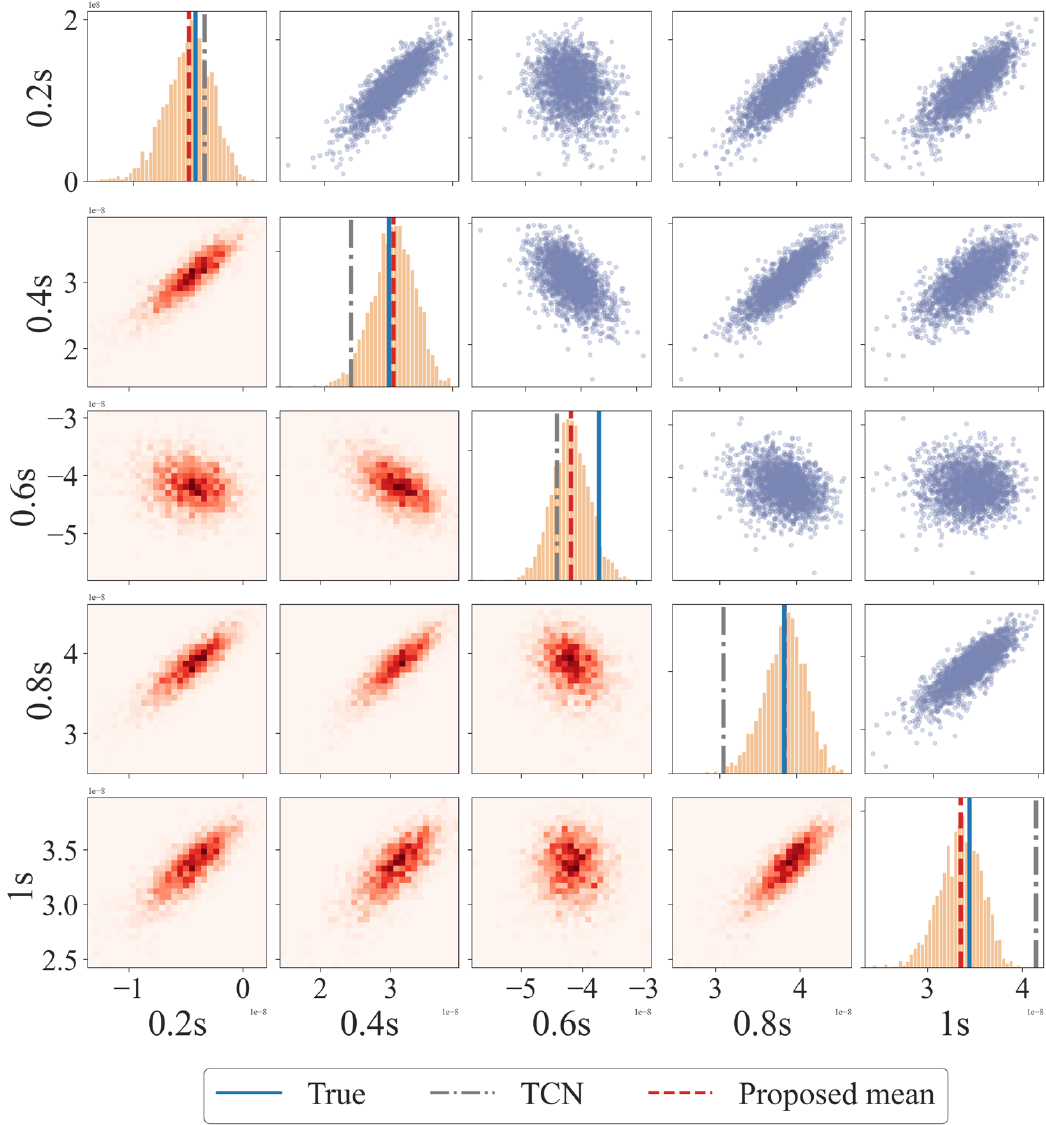}
        \caption{test \#2}
        \label{fig:dis_b}
    \end{subfigure}
    \caption{Probability distributions of predicted HF responses at selected time step for two test data examples.}
    \label{fig:comparison_dis}
\end{figure}

\Cref{fig:comparison_dis} presents the estimated probability distributions at selected time instants ($t = 0.2, 0.4, 0.6, 0.8,$ and $1.0$ s) for two HF test data pairs. In each subplot, the histograms depict the probability distributions estimated by the proposed method, the gray dot–dashed vertical lines correspond to the deterministic predictions of TCN, and the red dashed vertical lines indicate the predictive means of the proposed method. The results demonstrate the clear advantages of the proposed framework. In several time steps, the TCN predictions noticeably deviate from the ground truth, indicating the limitations of deterministic point estimation. By contrast, the proposed method consistently provides predictive means that are well aligned with the true responses while simultaneously delivering full probability distributions. These distributions capture the variability of the HF responses and thus enable uncertainty quantification, offering a more reliable and informative representation of the system dynamics compared to conventional ML approaches.

\section{Conclusion}\label{sec:conclusion}
In this study, we proposed a novel probabilistic MF surrogate modeling framework that integrates generative AI with transfer learning to address the challenge of HF data scarcity. The framework adopts a two-stage strategy: a generative surrogate is first pre-trained on abundant LF data, and the learned knowledge is then transferred via fine-tuning using limited HF data to construct a final MF model.
The backbone of the framework is a conditional surjective NF model. Unlike traditional bijective NF that preserves dimensionality and thus struggle with high-dimensional response data, our framework incorporates a surjective layer (via a funnel architecture) that enables dimensionality reduction while retaining likelihood-based training. This design allows the model to efficiently handle high-dimensional structural responses for given structural model parameters as inputs, providing not only accurate predictions but also uncertainty quantification.

The effectiveness of the framework was demonstrated for two benchmark examples: a rail-sleeper-ballast system and a reinforced concrete slab structure. In both applications, the proposed MF model exhibited strong predictive performance, accurately capturing the dynamics of high-dimensional time-series responses while providing reliable uncertainty estimates. Furthermore, an investigation on the effect of the amount of HF data used showed that fine-tuning the LF model consistently improves accuracy compared to training with HF data alone, underscoring the critical role of LF knowledge transfer when HF data is scarce.

The significance of this framework lies in its ability to combine generative modeling with transfer learning, enabling robust surrogate modeling under conditions of HF data scarcity. By explicitly quantifying uncertainty, our approach enhances the reliability of predictions, a feature that is essential for risk-informed decision-making in structural engineering. 
Future research will focus on extending this framework to more complex structural systems involving nonlinear and non-stationary dynamics. Additional directions include exploring adaptive multi-fidelity training strategies, incorporating physics-informed priors directly into the generative model, and applying the framework to real-world, large-scale structural health monitoring datasets.

\section*{Acknowledgments}
This material is based upon work supported by the U.S. Department of Energy (DOE), Office of Science, Office of Advanced Scientific Computing Research.
Pacific Northwest National Laboratory is operated by Battelle for the DOE under Contract DE-AC05-76RL01830.

\bibliographystyle{elsarticle-num}
\bibliography{reference}

\end{document}